\pdfoutput=1

\documentclass[11pt]{article}

\usepackage[final]{acl}

\usepackage{times}
\usepackage{latexsym}
\usepackage{multicol}
\usepackage{stfloats}

\usepackage[T1]{fontenc}

\usepackage[utf8]{inputenc}

\usepackage{microtype}

\usepackage{inconsolata}

\usepackage{graphicx}

\usepackage[utf8]{inputenc} 
\usepackage[T1]{fontenc}    
\usepackage{hyperref}       
\usepackage{url}            
\usepackage{booktabs}       
\usepackage{amsfonts}       
\usepackage{nicefrac}       
\usepackage{microtype}      
\usepackage{xcolor}         
\usepackage{bm}
\usepackage{amsmath}
\usepackage{amsthm}
\usepackage{mathtools}
\usepackage{wrapfig}
\usepackage{bm}
\usepackage{multirow}
\usepackage{amsfonts}
\usepackage{thm-restate}
\usepackage{tabularx} 
\usepackage{tablefootnote}

\newcommand{\R}{\mathbb{R}}
\newcommand{\E}{\mathbb{E}}

\newcommand{\cL}{\mathcal{L}}
\newcommand{\cO}{\mathcal{O}}
\newcommand{\mD}{\mathcal{D}}
 \newtheorem{lem}{{Lemma}}
 \newtheorem{assumption}{Assumption}

\newtheorem{thm}{{Theorem}}

\usepackage{algorithm} 
\usepackage{algorithmic}
\hypersetup{colorlinks, linkcolor=blue, citecolor=blue, urlcolor=magenta, linktocpage, plainpages=false}

\usepackage{caption}

\newcommand{\toedit}{{}}

\title{ScaleBiO: Scalable Bilevel Optimization for LLM Data Reweighting}

\author{\bf Rui Pan$^1$\footnotemark[1]\protect\phantom{\footnotesize 1},
~ Dylan Zhang$^1$\footnotemark[1]\protect\phantom{\footnotesize 1},
~ Hanning Zhang$^1$\footnotemark[1]\protect\phantom{\footnotesize 1},
~ Xingyuan Pan$^1$\thanks{\, Equal Contribution.}\protect\phantom{\footnotesize 1}, 
~ Minrui Xu$^2$\protect\phantom{\footnotesize 1},
\\
\bf  Jipeng Zhang$^{2}$\protect\phantom{\footnotesize 1}, ~  Renjie Pi$^2$, ~  Xiaoyu Wang$^2$\protect\phantom{\footnotesize 1},
Tong Zhang$^1$
\\
$^1$University of Illinois Urbana-Champaign \\
$^2$The Hong Kong University of Science and Technology\\
\texttt{\{ruip4, shizhuo2, hanning5, xp12\}@illinois.edu}
\\
\texttt{\{mxubh, jzhanggr, rpi, maxywang\}@ust.hk}\\\texttt{tongzhang@tongzhang-ml.org}
}

\begin{document}
\maketitle
\begin{abstract}
Bilevel optimization has shown its utility across various machine learning settings, yet most algorithms in practice require second-order information, making it challenging to scale them up. Only recently, a paradigm of first-order algorithms has emerged in the theoretical literature, capable of effectively addressing bilevel optimization problems. Nevertheless, the practical efficiency of this paradigm remains unverified, particularly in the context of large language models (LLMs). This paper introduces the first scalable instantiation of this paradigm called \textbf{ScaleBiO}, focusing on bilevel optimization for large-scale LLM data reweighting. By combining with a recently proposed memory-efficient training technique called LISA, our novel algorithm allows the paradigm to scale to $\sim$30B-sized LLMs on $8\times$H100 GPUs, marking the first successful application of bilevel optimization under practical scenarios for large-sized LLMs. Empirically, extensive experiments on data reweighting verify the effectiveness of ScaleBiO for different-scaled models, including Llama-3-8B, Gemma-2-9B, Qwen-2-7B, and Qwen-2.5-32B, where bilevel optimization succeeds in instruction-following and math reasoning tasks, outperforming several popular baselines, including uniform sampling, influence-aware data filtering, and reference-model-based sampling methods. Theoretically, ScaleBiO ensures the optimality of the learned data weights, along with a convergence guarantee matching the conventional first-order bilevel optimization paradigm on smooth and strongly convex objectives.

\end{abstract}

\section{Introduction}
Data quality plays a crucial role in the success of Large Language Models (LLMs)~\citep{gunasekar2023phi,yang2024qwen2-5,dubey2024llama3}. Among various techniques for improving data quality, data reweighting has gained increasing attention for advancing LLMs, particularly in areas such as enhancing fairness~\citep{roh2021sample,roh2020fairbatch,roh2023drfairness}, accelerating pre-training~\citep{xia2024sheared,xie2024doremi}, strengthening training robustness~\citep{jain2024datarobustness}, and boosting transfer learning~\citep{xia2024less}. It is widely acknowledged that data reweighting and filtering techniques can lead to significant improvements across a diverse range of tasks~\citep{gunasekar2023phi,xu2024magpie,hu2024fox,yang2024qwen2-5,liu2024deepseekv3}.

On the other hand, Bilevel Optimization (BO) has emerged as a prominent area of research for solving this data re-weighting task, which draws substantial attention due to its effectiveness in numerous machine learning applications, such as hyperparameter optimization~\citep{pmlr-v22-domke12,pmlr-v37-maclaurin15,iterative_der,lorraine2020optimizing}, meta-learning~\citep{andrychowicz2016learning,franceschi2018bilevel,rajeswaran2019meta} and reinforcement learning~\citep{konda1999actor,hong2020two}. In its standard formulation, bilevel optimization involves a two-level hierarchical structure with inner-outer dependence,
\begin{align}\label{P:bilevel}
    \min_{\lambda \in \Lambda} \quad & \quad \mathcal{L}(\lambda) = L_{1}(\lambda, w_{\ast}(\lambda)) \quad\notag \\
    \text{s.t.} \,\, \quad & \quad w_{\ast}(\lambda) = \arg\min_{w} L_2(\lambda, w).
\end{align}
For example, on data reweighting tasks, $\lambda$ are weights of different data sources, $w$ represents the trainable model parameters, $w_{\ast}(\lambda)$ means the optimal parameters trained on a weighted dataset, while outer function $L_1$ and inner function $L_2$ stand for validation and training losses, respectively.



Despite the inherent flexibility and applicability of bilevel optimization across a wide range of problems, its extensive utilization in large-scale problems has been relatively limited thus far. The primary obstacle hindering the scalability of bilevel optimization arises from the interdependence between the upper-level and lower-level problems. 
The natural gradient-based iterative method of solving Problem (\ref{P:bilevel}) is to compute (or estimate) the hyper-gradient
\begin{align}\label{equ:hyper:grad}
\frac{\partial \mathcal{L}(\lambda)}{\partial \lambda} = \frac{\partial L_1(w_{\ast}(\lambda), \lambda)}{\partial \lambda}  + \frac{\partial L_1(w_{\ast}, \lambda)}{\partial w_{\ast}} \frac{\partial w_{\ast}(\lambda)}{\partial \lambda},
\end{align}
where the main challenge lies in efficiently computing or approximating the derivative $\partial w_{\ast}(\lambda)/\partial \lambda$ in (\ref{equ:hyper:grad}). 
There is a line of research~\cite{pmlr-v22-domke12,pmlr-v48-pedregosa16,grazzi2020iteration,lorraine2020optimizing, iterative_der,shaban2019truncated,grazzi2020iteration, ghadimi2018approximation,hong2020two,yang2021provably,ji2021bilevel,chen2022single} have been tempted to address this challenge. However, these works all require the computation of Hessian, Jacobian, or their products with vectors, which can be computationally expensive and memory-intensive for large-scale problems. Recently, ~\citet{kwon2023fully} proposed a fully first-order method for stochastic bilevel optimization via only the first-order gradient oracle. This approach addresses the challenges associated with second-order computations and offers promising potential for stochastic bilevel optimization.


Despite these groundbreaking advancements in algorithms and theory, the practical performance of theoretically-optimal bilevel optimization algorithms in large-scale real-world settings has yet to be thoroughly investigated. Aiming to close this gap, this paper considers a practical scenario where LLMs are fine-tuned with different sources of datasets.
We identify a significant challenge in determining the optimal sampling weights for each data source.
For instance, \citet{wang2024far} have demonstrated that LLMs' task-specific performance degrades in the presence of certain training datasets. However, the inclusion and combination of various datasets should intuitively enhance the models' overall performance with proper sampling weights. This data-task misalignment poses a primary challenge in training LLMs with multiple data sources:
\begin{center}
    \textit{How to balance each data source in the training dataset to obtain optimal performance?}
\end{center}
Various methods have been proposed in attempting to address this challenge. However, they either rely on intuitive preset~\citep{zhou2024lima,muennighoff2022crosslingual,du2022glm,almazrouei2023falcon} or lacks theoretical guarantees~\citep{xia2024less,xie2024doremi,xia2024sheared}, leading to suboptimal sampling weights. To this end, we test theoretical-optimal bilevel optimization in data re-weighting tasks for LLMs, aiming to overcome the limitations of existing methods.
Our primary contributions are summarized as follows:
\begin{itemize}
  \item We propose the first scalable and theoretically-optimal instantiation of bilevel optimization on large-sized LLM training problems, which is capable of scaling to models with $\sim$30 billion parameters.
  \item We successfully bridge the gap between theoretical advancements in bilevel optimization and their application in data reweighting, allowing the optimal data weights to be learnable for large-scale LLMs.
  \item We provide both experimental and theoretical results to demonstrate the effectiveness of ScaleBiO. Empirically, ScaleBiO outperforms popular data filtering/reweighting baselines, including uniform sampling, LESS~\citep{xia2024less}, and RHO-LOSS~\citep{mindermann2022rholoss}, surpassing them by a non-trivial margin of $1\%-9\%$ in GSM8K~\citep{cobbe2021gsm8k} and MATH~\citep{hendrycks2021math}. This superiority of ScaleBiO also holds in instruction following tasks. Theoretically, ScaleBiO's convergence guarantee matches the results of~\citet{kwon2023fully} on smooth and strongly convex objectives.
\end{itemize}

\section{Related Work}



\begin{table*}[th]
\small
	\centering

	\renewcommand    
	\arraystretch{1.3}
            \setlength{\tabcolsep}{1.5mm}
\begin{tabularx}{\linewidth}{XXlrr}
\toprule
Method  & Description   & Task & Model & Size \\ \midrule
RMD~\citep{bengio2000gradient} & 2-nd order, deterministic &   hyperparameter optimization & Linear & $<$1M \\
CG~\citep{grazzi2020iteration}     &  2-nd order, deterministic & equilibrium models & CNN &  $<$1M      \\
stocBiO~\citep{ji2021bilevel}  & 2-nd order, stochastic & meta learning &  CNN  &    $<$1M    \\
FdeHBO~\citep{yang2024achieving}& 1-st order, stochastic&hyper-representation &LeNet &$<$1M \\
BOME~\citep{liu2022bome}& 1-st order, stochastic & data hyper-cleaning & Linear & $<$1M \\
SOBA~\citep{dagreou2022framework} & 2-nd order, stochastic & data reweighting & Transformers &  7M \\
PZOBO~\citep{Sow_2022_zeroth}& 1-st order, stochastic &few-shot meta-learning&ResNet&12M \\
SAMA~\citep{choe2023sama} & 2-nd order, stochastic & noisy fine-tuning & BERT & 110M \\
BFTSS~\citep{somayajula2023bi} & 1-st order, stochastic  & task-dependent structure learning &  BERT & 336M \\
(FG)$^2$U~\citep{shen2024fg2u} & 1-st order, stochastic & online adaptation & GPT-2-XL & 1.5B \\
\midrule
ScaleBiO (Ours)     &  1-st order, stochastic  & data reweighting  &     Qwen-2.5-32B & \textbf{32B}       \\
 \bottomrule
\end{tabularx}
\caption{In this table, we compare the maximal model size implemented in their original paper, where 'M' stands for million and 'B' stands for billion. We also summarize their methods in Description and report the task they tested.}
\label{tab:bilevel_model_size_comparison}
\end{table*}

\subsection{Bilevel Optimization}

Traditional bilevel optimization algorithms are majorly categorized into two classes: 1) approximate implicit differentiable (AID) methods~\cite{pmlr-v22-domke12,pmlr-v48-pedregosa16,grazzi2020iteration,lorraine2020optimizing}, or 2) iterative differentiable  (ITD) methods~\cite{pmlr-v22-domke12,pmlr-v37-maclaurin15,iterative_der,shaban2019truncated,grazzi2020iteration}. Both approaches follow a two-loops manner and require huge computational cost for large-scale problems. To reduce the cost, attempts in stochastic bilevel optimization have been made~\citep{ghadimi2018approximation,hong2020two,ji2021bilevel,chen2022single,khanduri2021near}, which significantly improve the efficiency of traditional methods, but still lack practicality for large-scale settings due to the requirements of second-order information, such as Jacobian- and Hessian-vector products for estimating the hyper-gradient.~\citet{Sow_2022_zeroth,yang2024achieving} attempt to approximate the Jacobian matrix $\nabla y^{\ast}(x)$ in~\eqref{equ:hyper:grad} by finite differences, but the finite-different estimation can be sensitive to the selection of the smoothing constant and may suffer from some numerical issues in practice~\cite{jorge2006numerical}.



Recently, a new paradigm of fully first-order penalty-based methods has been introduced, which reformulate the inner-level problem into the optimality constraint~\cite{liu2022bome,kwon2023fully,chen2023nearoptimal}. 
\citet{liu2022bome} first find the hypergradient only involving first-order information, while the method only applies to deterministic functions. \citet{kwon2023fully} introduced a first-order gradient-based approach that avoids the estimations of Hessian or Jacobian. This method is easily adapted and extended to stochastic bilevel optimization settings. \citet{chen2023nearoptimal} provided the near-optimal sample complexity, which improves the theoretical result of~\cite{kwon2023fully} in the deterministic bilevel optimization. These results verify the effectiveness of the proposed paradigm in theory, yet its practical applications in large-scale LLM settings remain unexplored.

On the practical side, bilevel optimization has been explored in various NLP tasks. \citet{somayajula2023bi} use bilevel optimization to learn the task-dependent similarity structure. \toedit{Although their approach demonstrates effectiveness on BERT models~\citep{devlin2018bert}, the finite difference approximation suffers from high error and therefore lacks the scalability in LLMs with billions of parameters.} \citet{grangier2024bilevel} adopt SOBA~\citep{dagreou2022framework} to modify the training data distributions for language modeling under domain shift. However, the algorithm still requires gradient approximation and Hessian-vector products, posing challenges to scalability and engineering for large-scale problems. We summarize typical bilevel algorithms and their model sizes in Table~\ref{tab:bilevel_model_size_comparison}, where to the best of our knowledge, no approach listed in the table has been successfully applied to over 3B-sized LLM models.

\subsection{Data Reweighting}
The proportion of training data sources significantly affects the performance of large language models~\citep{du2022glam,xie2023data}. To this end, various methods have been proposed to reweight data sources for optimal training data mixture. For example, ~\citet{mindermann2022prioritized} utilizes the loss gap between a trained model and a base model to identify learnable data samples, assigning them higher weights on the fly.~\citet{thakkar2023self} propose to use a self-influence score to guide the reweighting in mini-batch during pre-training.~\citet{xia2024sheared} leverages reference losses on validation sets and adjusts the weights dynamically, adding minimal
overhead to standard training. DoReMi~\citep{xie2024doremi} applies distributionally robust optimization (DRO) to tuning the domain weights without knowledge of downstream tasks. Nevertheless, none of the aforementioned methods ensures the optimality of the learned data weights, let alone scalable experiments on over 30B-sized models.

\section{Methods}
In this section, we elaborate on our ScaleBiO method for finding the optimal sampling weights when training large-scale LLMs. We first formulate this problem as a bilevel optimization problem in Section \ref{sec:prob_form} and then develop an efficient training method for our formulation in Section \ref{sec:prob_opt}.

\subsection{Problem Formulation}
\label{sec:prob_form}
Suppose that $m$ data sources are available for training, e.g. Alpaca~\citep{alpaca}, FLAN~\citep{wei2021flan}, and ShareGPT~\citep{vicuna2023}, where each source $S_i$ is a set of $n_i$ examples $S_i=\{a_1^i, a_2^i, \dots, a_{n_i}^i\}$. The desired dataset mixture can be obtained by assigning each data source $S_i$ a sampling weight $p_i$ that satisfies $\sum_{i=1}^mp_i=1$.

Accordingly, each data source $S_i$ contributes $p_i\vert\mD_{\text{trn}}\vert$ samples to the training dataset $\mD_{\text{trn}}$, where the sampling weights can be optimized to minimize the model's loss on validation set $\mD_{\text{val}}$. This leads to the following bilevel optimization problem:
\begin{align*}
    \min_{p\in\Lambda} &\,\, L_{\text{val}}(w^*(p)) \notag \\ 
    \mathrm{s.t.}  & \,\,
          w^*(p)=\arg\min_w\sum_{i=1}^m \frac{p_i}{n_i}\sum_{j=1}^{n_i}L_{\text{trn}}(w, a_j^i)
\end{align*}
where $w$ denotes the parameters of LLM, $\{p_i\}$ is the probability distribution over $m$ data sources, $L_{\text{val}}$ and $L_{\text{trn}}$ respectively denote the language modeling loss on $D_{\text{val}}$ and $D_{\text{trn}}$.
To ensure non-negativity of the sampling weights $\{p_i\}_{i=1}^m$, an additional trainable variable $\lambda\in\R^m$ is introduced to represent $p_i = e^{\lambda_i} / \sum_{j=1}^m e^{\lambda_j}$.

 \begin{algorithm*}[t]
\caption{ScaleBiO for high-dimensional and large-scale minimax problems}
\label{alg:minmax:sgd}
\begin{algorithmic}[1]
\STATE {\bfseries Input:} step-sizes $\left\lbrace \eta_{u}, \eta_{\omega}, \eta_{\lambda}\right\rbrace$, penalty $\alpha$, and initialization $\lambda_0$, $u_0$,  $w_0$
\FOR{$k=0: K-1$}
\STATE{Uniformly and independently select two $j_k, r_k$ block coordinates from $\left\lbrace 1,2,\cdots, J\right\rbrace$, respectively}
\STATE{Generating i.i.d. samples  $\left\lbrace D_{\text{tr}}^k, D_{\text{val}}^k \right\rbrace$ from training dataset $D_{\text{tr}}$ and validation dataset $D_{\text{val}}$}
\STATE{$u_{k+1}^{j_k} = u_k^{j_k} - \alpha\eta_u \nabla_{j_k}L_2(\lambda_k, u_k; D_{\text{tr}}^k) $
}
\STATE{$u_{k+1} = u_k + U_{j_k}(u_{k+1}^{j_k}-u_{k}^{j_k})  \text{\quad \quad \quad \quad \quad \quad \quad \quad \textcolor{gray}{$\triangleright$ Map the permuted block parameters back}}$} 
\STATE{$w_{k+1}^{r_k} = w_k^{r_k} -\eta_w \left(\nabla_{r_k} L_1(\lambda_k, w_k; D_{\text{val}}^k) + \alpha \nabla_{r_k} L_2(\lambda_k, w_k; D_{\text{tr}}^k) \right)$
}
\STATE{$w_{k+1} = w_k + W_{r_k}(w_{k+1}^{r_k}-w_{k}^{r_k}) \text{\quad \quad \quad \quad \quad \quad \quad  \textcolor{gray}{$\triangleright$ Map the permuted block parameters back}}$} 
\STATE{$\lambda_{k+1} = \lambda_k - \eta_{\lambda} \left(\nabla L_1(\lambda_k, w_k; D_{\text{val}}^k) + \alpha \bigl(\nabla L_2(\lambda_k, w_k; D_{\text{tr}}^k) -  \nabla L_2(\lambda_k, u_k; D_{\text{tr}}^k)\bigl)\right)$
}
\ENDFOR
\STATE{{\bf Output:} $(\lambda_K, w_K, u_K)$}
\end{algorithmic}
\end{algorithm*}

\subsection{Fully First-order Hypergradient Method}
\label{sec:prob_opt}

Recent advancements in the theoretical literature of bilevel optimization allow scalable methods to be developed. The main idea is actually quite similar to merging digits in radix sort. The first step is to decouple two ``digit'' terms and view the inner-level problem in~\eqref{P:bilevel} as a higher-order digit,
\begin{align}\label{P:bilevel:constraint}
    \min_{\lambda \in \Lambda, w} & \quad  L_{1}(\lambda, w) \notag \\
    \text{s.t.}  &  \quad L_2(\lambda, w)  - \min_{u} L_2(\lambda, u) = 0.
\end{align}
Here the auxiliary variable $u$ is introduced to detach the inner-outer dependency, which transforms the inner problem $w_{\ast}(\lambda) = \arg\min_{w} L_2(\lambda, w)$ to be the constraint $L_2(\lambda, w)  - \min_{u} L_2(\lambda, u) = 0$. By prioritizing the ``high-order'' constraint term of \eqref{P:bilevel:constraint} with multiplier $\alpha > 0$, the minimax formulation in~\citet{kwon2023fully, lu2023first} is recovered:
\begin{align}\label{P:minimax}
\min_{\lambda \in \Lambda, w} \max_{u} \,\,\mathcal{L}^{\alpha}(\lambda, w, u).
\end{align}
where 
\begin{align*}
\mathcal{L}^{\alpha}(\lambda, w, u)=  L_1(\lambda, w) + \alpha \left(L_2(\lambda, w) - L_2(\lambda, u) \right)
\end{align*}
In this way, the approximation of both inner constraint and outer
optimum can be obtained during the same optimization process, and $\alpha$ controls the priority. When $\alpha \to \infty$, the bilevel problem~\eqref{P:bilevel} is equivalent to the minimax problem~\eqref{P:minimax} under certain smoothness assumptions.

To precisely describe the optimality of the minimax problem with the stationarity of the bilevel problem, the following notations in \eqref{P:minimax} are overloaded and defined as
\begin{align}\label{P:minimax:u}
    \Phi^{\alpha}(\lambda, w) & := \max_{u} \mathcal{L}^{\alpha}(\lambda, w, u);  \\
    u_{\ast}(\lambda) &:= \arg\max_{u} \mathcal{L}^{\alpha}(\lambda, w, u); \\
    \Gamma^{\alpha}(\lambda) & := \min_{w} \Phi^{\alpha}(\lambda, w);  \\
    w_{\ast}^{\alpha}(\lambda) & = \arg\min_{w} \Phi^{\alpha}(\lambda, w).
\end{align}
Additionally, the following assumptions are made for the proposed minimax problem throughout this paper.
\begin{assumption}\label{assumpt:v2}
Suppose that
\begin{itemize}
    \item[(1)] $L_1(\lambda, w)$ is twice continuously differentiable, $\ell_{10}$-Lipschitz continuous in $w$; $\ell_{11}$-gradient Lipschitz.
    \item[(2)]   $L_2(\lambda, w)$ is $\ell_{21}$-gradient Lipschitz, $\ell_{22}$-Hessian Lipschitz, and $\mu_2$-strongly convex in $w$.
\end{itemize}
\end{assumption}
\begin{restatable}{lem}{thmequivstronger} \label{lem:equiv}
Under Assumption \ref{assumpt:v2}, if $\alpha > 2\ell_{11}/\mu_2$, we have
\begin{align}
|\mathcal{L}(\lambda) - \Gamma^{\alpha}(\lambda)| & \leq  \mathcal{O}\left(\frac{1}{\alpha}\right)  \label{inequ:zero:dif} \\
 \left\| \nabla \mathcal{L}(\lambda) - \nabla \Gamma^{\alpha}(\lambda)\right\| & \leq  \mathcal{O}\left(\frac{1}{\alpha}\right)\label{inequ:one:dif} \\
 \left\|\nabla^2 \Gamma^{\alpha}(\lambda) \right\| & \leq \mathcal{O}(\kappa^3) \label{inequ:sec:dif}
\end{align}
where the condition number $\kappa$ is defined by $\max\left\lbrace \ell_{10}, \ell_{11}, \ell_{21}, \ell_{22} \right\rbrace / \mu_2$.
\end{restatable}
Under Assumption~\ref{assumpt:v2}, as indicated by Lemma~\ref{lem:equiv}, if $\alpha$ goes to infinity, the stationary point of the minimax problem~\eqref{P:minimax} is also a stationary point of the bilevel problem~\eqref{P:bilevel}. Intuitively, it is like finding a way to the same peak of the mountain with distinct paths, where bilevel optimization suggests a winding road, and minimax utilizes a helicopter.

\subsection{Proposed Algorithm}

To solve this reformulated large-scale min-max problem, we introduce ScaleBiO in Algorithm~\ref{alg:minmax:sgd}, a single-loop framework that is capable of scaling up to 30B-sized models. To further reduce memory consumption, randomized block coordinate methods are employed to update the inner variables $u, w$~\citep{nesterov2012efficiency, pan2024lisa}, where $U_j, W_j \in \mathbb{R}^{d\times d_j}$ denotes the block matrices that map the permutation of parameters back to model weights. The optimizer choice varies depending on the backbone model, where Adam or AdamW~\citep{Adam, loshchilov2017adamw} is much preferable for LLMs. The penalty coefficient $\alpha$ is predefined with a large factor that ensures the min-max solution is a good approximation of the original bilevel problem.



\subsection{Theoretical Results}
In this part, we provide a convergence analysis of Algorithm~\ref{alg:minmax:sgd}, explaining how fast the algorithm can reach a desired stationary point. Before showing the details of theoretical results, we introduce the notations for partitions. Let $\left\lbrace x^{1}, x^{2}, \cdots, x^{J}\right\rbrace$ with $x^{j} \in \R^{d_j \times 1}$ be $J$ non-overlapping blocks of $x$. Let the matrix $U_{j} \in \R^{d \times d_j}$ be $d_j$ columns of a $d \times d$ permutation matrix $U$ corresponding to $j$ block coordinates in $x$. For any partition of $x$ and $U$,
\begin{align}
x = \sum_{j=1}^{J} U_jx^j, \quad x_j = U_j^{T}x.
\end{align}

The essential lemmas are available in Appendix~\ref{sec:appendix:lem} to show the theoretical properties of minimax objective $\cL^{\alpha}$ in~\eqref{P:minimax}, as well as its optimizers $u_{\ast}$ and $w_{\ast}^{\alpha}$. Lemma~\ref{lem:equiv} provides clear evidence that $\Gamma^{\alpha}(\lambda)$ is smooth with parameter $\ell_{\Gamma} = \mathcal{O}(\kappa^3)$ which is independent on the multiplier $\alpha$.
\begin{thm}\label{thm:bilevel:complexity}
Suppose that Assumptions~\ref{assumpt:v2} holds and the parameter $\alpha$ and step-sizes $\eta_u, \eta_{w}, \eta_{\lambda}$ are properly chosen such that 
\begin{align*}
\alpha = K^{1/7}, \eta_{u} = \eta_{w} = \frac{\eta_0}{K^{4/7}}, \eta_{\lambda} = \frac{\eta_0^{\lambda}}{K^{5/7}}.
\end{align*}
Consider Algorithm~\ref{alg:minmax:sgd}, if $\alpha \geq \ell_{11}/\mu_2$, for $\eta_0^{\lambda}\leq 1/(8\ell_{\Gamma})$, $\eta_0 \leq 8J/\mu_2$ and $\eta_0/\eta_{0}^{\lambda} \geq 6\sqrt{2}\kappa^2J$, then
\begin{align}
\E\left[\left\|\nabla \cL(\tilde{\lambda})  \right\|^2 \right] \leq \mathcal{O}\left( \frac{1}{K^{2/7}}\right)
\end{align}
where $\tilde{\lambda}$ is uniformly chosen from $\left\lbrace \lambda_k\right\rbrace_{k=1}^{K}$. 
\end{thm}
When considering the batch size $B = \mathcal{O}(1)$, the complexity of finding an $\epsilon$-stationary point of Algorithm~\ref{alg:minmax:sgd} is $\mathcal{O}(\epsilon^{-7})$, which matches that of \citep{kwon2023fully}. The proof of Theorem~\ref{thm:bilevel:complexity} is provided in Appendix~\ref{sec:proof:thm}.

\section{Experiments}
To verify the effectiveness of ScaleBiO, two types of experiment are conducted: (1) \textit{Small Scale Experiments} in Section~\ref{sec:poc_exp}, which offers intuitions for understanding ScaleBiO's theoretical properties in toy settings, and (2) \textit{Real-World Application Experiments} in Section \ref{sec:rwa_exp} that validate its scalability in larger-sized models on instruction-following and mathematical reasoning tasks.

\begin{figure}
\centering
  \includegraphics[width=1.0\linewidth]{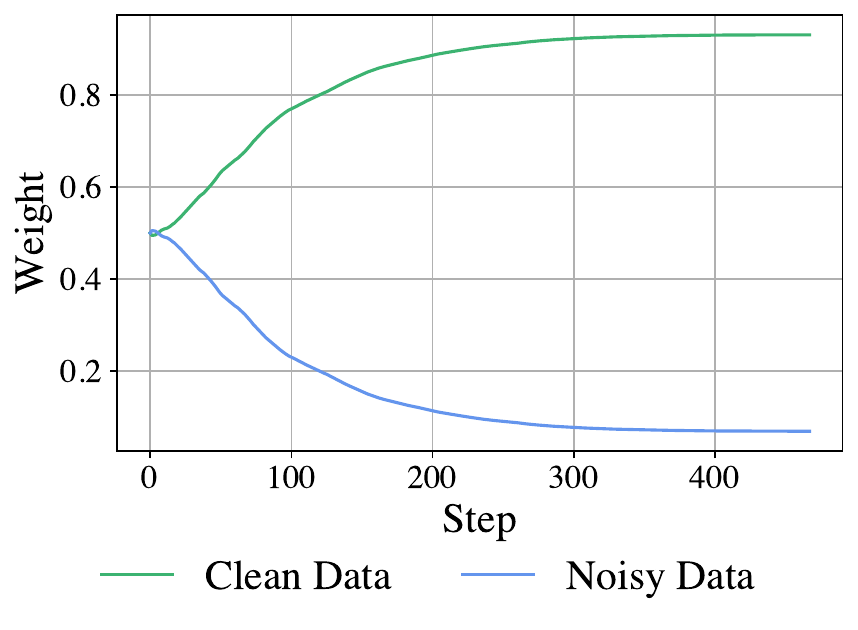}
  \caption{\textbf{Data denoising with GPT-2}: weights for noisy data and clean data.}
  \label{fig:denoise}
\end{figure}

\subsection{Small Scale Experiments}
\label{sec:poc_exp}
To understand the properties of ScaleBiO, experiments with GPT-2 (124M) are conducted on three tasks with synthetic datasets: data denoising, multilingual training, and instruction-following fine-tuning. Full details are available in Appendix~\ref{sec:appendix:small_scale_exp_setup}.



\subsubsection{Data Denoising}
In this experiment, ScaleBiO's data denoising ability is tested, where noisy samples are expected to be assigned with zero weights. The validation dataset, denoted as $ \mD_\text{val}$, comprises 1000 clean samples randomly selected from the Alpaca dataset~\cite{alpaca}. The training dataset, $ \mD_\text{trn}$, is derived from two distinct sources: the first includes 1000 clean samples also from Alpaca, while the second incorporates 9000 samples from Alpaca that have been artificially corrupted with synthetic noise, where the outputs are replaced with ".".

Figure~\ref{fig:denoise} demonstrates that our approach has a robust capability to mitigate the influence of harmful data sources via automatic data denoising, where ScaleBiO assigns minimal weight to noisy data sources, effectively filtering the irrelevant samples.

\subsubsection{Multilingual Training} 
It is also intriguing to check if ScaleBiO can recover optimal sampling weights for more general distributions. To this end, the multilingual training experiments are introduced, where the validation data $\mD_{\text{val}}$ comprises 600 random samples from Alpaca-GPT4-ZH~\cite{peng2023instruction} and 400 random samples from Alpaca-GPT4-EN~\cite{peng2023instruction}. Hence, the underlying optimal weight is 6:4. In contrast, the training set $\mD_{\text{trn}}$ has a 1:1 mix ratio, which consists of 40,000/40,000 random examples from Alpaca-GPT4-EN and Alpaca-GPT4-ZH, respectively.

\begin{figure}
\centering
  \includegraphics[width=1.0\linewidth]{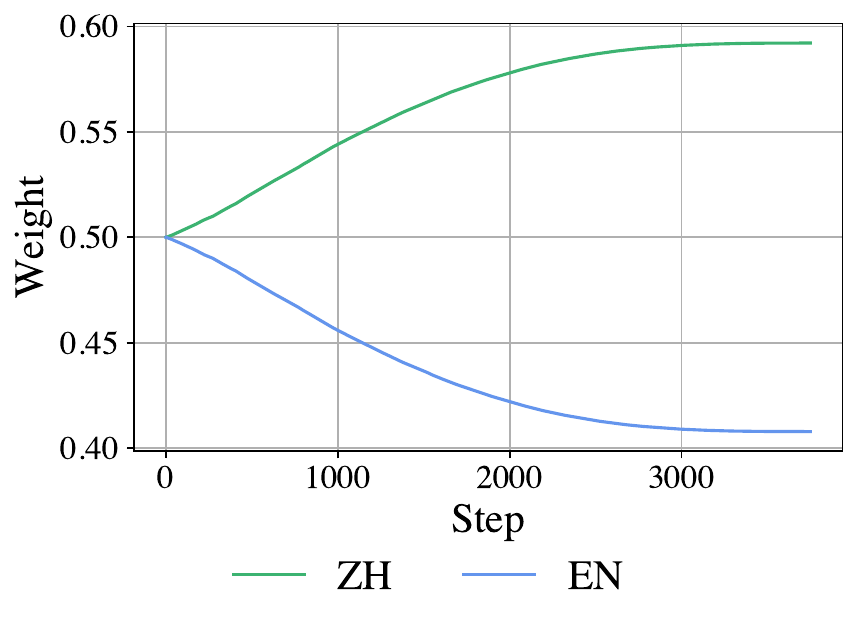} 
  \caption{\textbf{Multilingual reweighting with GPT-2}: weights of Chinese and English. Training set: 1:1; Validation set: 6:4.}
  \label{fig:multilingual}
\end{figure}

As shown in Figure~\ref{fig:multilingual}, ScaleBiO nearly replicates the optimal 6:4 ratio after reweighting the training data. This serves as another concrete proof that ScaleBiO is capable of adapting training data weights optimally to downstream validation datasets.

\begin{figure}[b!]
\centering
  \includegraphics[width=1.0\linewidth]{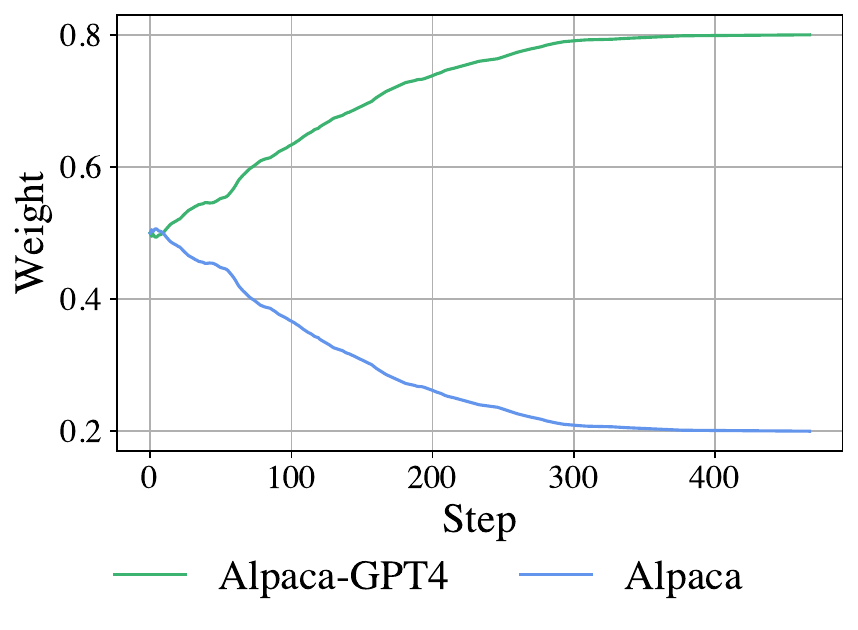} 
  \caption{\textbf{Instruction Following with GPT-2}: weights for Alpaca-GPT4 and Alpaca.}
  \label{fig:ift}
\end{figure}



\subsubsection{Instruction Following} In instruction-following fine-tuning tasks, there is a fundamental tradeoff between diversity and quality. To verify if ScaleBiO can deduce these implicit weights of low- and high-quality datasets, experiments are conducted on instruction-following tasks with GPT-2, where Alpaca and Alpaca-GPT4~\cite{peng2023instruction} are employed. Here Alpaca-GPT4 shares the same instructions and input as Alpaca, whose high quality is distinguished by its outputs generated from a more sophisticated model GPT-4~\citep{achiam2023gpt4}. The validation data for bilevel optimization $\mD_{\text{val}}$ consists of 1000 random samples from Alpaca-GPT4, while the training data $\mD_{\text{trn}}$ consists of 2 separate parts: 1000 random samples from Alpaca-GPT4 and 9000 random samples from Alpaca.

As shown in Figure \ref{fig:ift}, although Alpaca-GPT4 accounts for only a small proportion of the training data (10\%), it is highlighted by ScaleBiO, revealing that it can effectively up-weights the high-quality data source, leading to improved model outcomes.

\subsection{Real-World Application Experiments}
\label{sec:rwa_exp}

In this section, ScaleBiO is tested in real-world data reweighting applications, including instruction following and mathematical reasoning tasks, which demonstrates its scalability and empirical benefits in practice.

\subsubsection{Instruction Following}
\label{sec:exp:instruction_following}

\begin{table*}[t]\small
  \centering
  \begin{tabular}{l|rrr}
    \toprule
    \multirow{2}{*}{Method} & \multicolumn{3}{c}{Model}
    \\
    & Llama-3-8B & Qwen-2-7B & Gemma-2-9B
    \\ \midrule
    SOBA~\citep{dagreou2022framework} & OOM & OOM & OOM
    \\
    Uniform Weighting & 6.11 & 6.66 & 5.31
    \\
    LESS~\citep{xia2024less} & 6.06 & 7.18 & 7.20
    \\
    RHO-LOSS~\citep{mindermann2022rholoss} & 6.89 & 7.34 & 7.38
    \\ \midrule
    ScaleBiO & \textbf{7.12} & \textbf{7.76} & \textbf{7.51}
    \\
    \bottomrule
\end{tabular}
\caption{\textbf{Instruction Following}. Here all methods are evaluated in MT-Bench~\citep{zheng2023judging} with GPT-4o LLM judge, where scores range from 0 to 10. OOM stands for out of memory.}
\label{tab:main_instruction_following}
\end{table*}

In the instruction following setting, ScaleBiO is validated under the real-world scenario where the data collection is conducted in a non-filtered fashion, e.g. datasets of weak correlations with the downstream task may be included.

\paragraph{Setup} Three $\sim$7B-sized LLMs, including Llama-3-8B~\citep{dubey2024llama3}, Qwen-2-7B~\citep{yang2024qwen2}, and Gemma-2-9B~\citep{team2024gemma2} are evaluated in the widely adopted benchmark of MT-Bench~\citep{zheng2023judging}, where a GPT-4o judge~\citep{hurst2024gpt4o} is employed to score the generated responses of each model on 80 high-quality multi-turn questions. Different aspects of the model, such as writing, role play, and STEM, are scored by the GPT-4o judge and averaged in the final MT-Bench score. 

The training portfolio comprises $\sim$4.2M total samples from 18 different sources, as detailed in Table~\ref{tab:appendix:data_info_inst_following} in Appendix~\ref{sec:appendix:large_scale_exp_setup}. All datasets are collected in a task-agnostic fashion, where datasets necessary for general instruction following tasks, but have weak correlations to MT-Bench are also included. One typical example of such datasets is multi-lingual conversations.

To form the training set, all data reweighting methods are required to assign weights to 18 sources and extract 10K samples from the 4.2M portfolio. The target model will be trained on the training set and evaluated to produce the final MT-Bench score.

\paragraph{Results} As shown in Table~\ref{tab:main_instruction_following}, ScaleBiO is the only bilevel algorithm capable of yielding meaningful weights across data sources. On top of that, SaleBiO outperforms popular influence-aware data filtering method LESS~\citep{xia2024less} and reference-model-based data reweighting approach RHO-LOSS~\citep{mindermann2022rholoss}, both are considered strong non-bilevel baselines in the data reweighting literature.

\subsubsection{Mathematical Reasoning}
\label{sec:exp:math_reasoning}

To further demonstrate ScaleBiO's data reweighting ability under scenarios with refined dataset sources, a training portfolio similar to~\citet{dong2024rlhflow} is adopted for mathematical reasoning, where datasets detailed in Table~\ref{tab:main:dataset_info} are proven to be conducive to downstream math tasks. Here coding and instruction following datasets are considered necessary, which allow the LLM to learn minimum reasoning and instruction following abilities for answering mathematical questions.

\begin{table}[h!]\small
\centering
\resizebox{.5\textwidth}{!}{
\begin{tabular}{@{}lr@{}}
\toprule
\toprule
\textbf{Dataset} & \textbf{\#Samples} \\ \midrule
\texttt{hkust-nlp/dart-math-uniform} & 591K \\
\texttt{Open-Orca/SlimOrca} & 518K \\
\texttt{openbmb/UltraInteract\_sft} & 289K \\
\texttt{TIGER-Lab/MathInstruct} & 262K \\
\texttt{microsoft/orca-math-word-problems-200k} & 200K \\
\texttt{WizardLMTeam/WizardLM\_evol\_instruct\_V2\_196k} & 196K  \\
\texttt{ise-uiuc/Magicoder-Evol-Instruct-110K} & 110K \\
\texttt{anon8231489123/ShareGPT\_Vicuna\_unfiltered} & 94K \\
\texttt{teknium/GPTeacher-General-Instruct} & 89K \\
\texttt{teknium/GPT4-LLM-Cleaned} & 55K \\
\midrule
\textbf{Total}                     & \textbf{2.4M} \\ \bottomrule \bottomrule
\end{tabular}
}
\caption{Dataset for \textbf{Mathematical Reasoning}.}
\label{tab:main:dataset_info}
\end{table}

\paragraph{Setup} Similar to Section~\ref{sec:exp:instruction_following}, three models of Llama-3-8B, Qwen-2-7B and Gemma-2-9B are employed. For evaluation, the standard math benchmark of GSM8K~\citep{cobbe2021gsm8k} and MATH~\citep{hendrycks2021math} are utilized. The reweighting methods are expected to extract 20K samples from 
 the given 10 sources to form the training set, with the target model fine-tuned on the set and evaluated to produce the final accuracy.

\begin{table*}[t]\small
  \centering
  \begin{tabular}{l|rrrrrr}
    \toprule
    \multirow{2}{*}{Method} & \multicolumn{3}{c}{GSM8K~\citep{cobbe2021gsm8k}} & \multicolumn{3}{c}{MATH~\citep{hendrycks2021math}}
    \\
    & Llama-3-8B & Qwen-2-7B & Gemma-2-9B & Llama-3-8B & Qwen-2-7B & Gemma-2-9B
    \\ \midrule
    SOBA & OOM & OOM & OOM & OOM & OOM & OOM
    \\
    Uniform Weighting & 53.6 & 65.0 & 56.3 & 14.2 & 36.7 & 24.8
    \\
    RHO-LOSS & 53.8 & 70.7 & 56.9 & 13.6 & 38.8 & 25.0
    \\
    LESS & 52.5 & 71.6 & 57.9 & 14.0 & 38.9 & 28.3
    \\ \midrule
    ScaleBiO & \textbf{56.2} & \textbf{74.1} & \textbf{59.4} & \textbf{15.1} & \textbf{41.7} & \textbf{30.0}
    \\
    \bottomrule
\end{tabular}
\caption{\textbf{Mathematical Reasoning}. Here all metrics are accuracies ranging from 0 to 100. OOM stands for out of memory.}
\label{tab:main_math_reasoning}
\end{table*}

\paragraph{Results} As shown in Table~\ref{tab:main_math_reasoning}, ScaleBiO consistently outperforms all baselines across different models and benchmarks, by a non-trivial margin of 1\%-9\%, which demonstrates ScaleBiO's superiority in reweighting task-oriented datasets.

\begin{table}[h]\small
  \centering
  \begin{tabular}{l|rr}
    \toprule
    \multirow{1}{*}{Method} & GSM8K & MATH
    \\ \midrule
    LESS & OOM & OOM
    \\
    RHO-LOSS & OOM & OOM
    \\
    Uniform Weighting & 78.1 & 54.0
    \\ \midrule
    ScaleBiO & \textbf{87.1} & \textbf{59.8}
    \\
    \bottomrule
\end{tabular}
\caption{\textbf{Large-Scale Mathematical Reasoning} on Qwen-2.5-32B~\citep{yang2024qwen2-5}. Here all metrics are accuracies ranging from 0 to 100. OOM stands for out of memory.}
\label{tab:main:math_large_scale}
\end{table}

To further validate ScaleBiO's scalability in even larger-sized LLMs, Qwen-2.5-32B~\citep{yang2024qwen2-5} is adopted in the same setting. As presented in Table~\ref{tab:main:math_large_scale}, ScaleBiO is the only data reweighting implementation capable of scaling up to this size. Here LESS and RHO-LOSS both run out of GPU memories due to their non-scalable implementation or requirement for extra reference models. In contrast, ScaleBiO has the same space complexity as full parameter fine-tuning, allowing it to be applicable in any single-node training scenarios.

\section{Discussion}

\paragraph{Existence of Optimal Datasets?} As ScaleBiO is capable of learning optimal task-orient data weights for different models, it serves as a great tool to inspect data weight transferability across different models. As it is unsurprising to find that Llama-3-8B-learned data weights can be transferred to Llama-3-70B and still yield certain improvement (Table~\ref{tab:7B-to-70B}), it is more intriguing to observe that the learned data weights vary significantly across different model families, as shown in Table~\ref{tab:appendix:model-family-weight} of Appendix~\ref{appendix:model_family_weights}.

\begin{table}[h]\small
	\renewcommand    
	\arraystretch{1.3}
	\centering
            \setlength{\tabcolsep}{1.5mm}
\begin{tabular}{@{}l | l c@{}}
\toprule
Model & \multicolumn{2}{r}{MT-Bench score} \\ \midrule

\multirow{2}{*}{Llama-3-8B $\rightarrow$ Llama-3-70B} & \textit{Uniform Weighting}  & 7.85  \\
                            & \textit{ScaleBiO}    &  \textbf{8.05}  \\
\bottomrule
\end{tabular}
\caption{MT-Bench results of Llama-3-70B with transfer trained weights from Llama-3-8B.}
\label{tab:7B-to-70B}
\end{table}

It is worth noticing that the weight difference is much smaller inside the same model family. This phenomenon is conjectured to stem from the difference in LLMs' pre-training dataset distributions, where the strengths of different models may vary from each other and need distinct datasets to adapt to the same downstream task. In that case, optimal dataset weights across models would be impossible for small-sized dataset settings, leaving model-dependent reweighting be the only choice.

\section{Conclusion}
In this paper, we propose ScaleBiO, the first bilevel optimization instantiation that is capable of scaling to over 30B-sized LLMs on data reweighting tasks. Theoretically, ScaleBiO ensures optimality of the learned data weights and enjoys the same convergence guarantees as conventional first-order penalty-based bilevel optimization algorithms on smooth and strongly convex objectives. Empirically, ScaleBiO enables data reweighting on $>$30B-sized models, bringing forth an efficient data filtering and selection pipeline for improving model performance on various downstream tasks.

\section*{Limitations}

The proposed algorithm of ScaleBiO has yet to be verified in large-scale pre-training settings, where a huge amount of computation resources are required for conducting such experiments. We hope the success of ScaleBiO in large-scale fine-tuning settings can be the first step towards this direction.

The potential risks of ScaleBiO are the same as other data reweighting techniques, where optimizing the sampling weights on a single loss metric may lead to models that neglect other aspects, such as safety or ethics. In that case, multi-objective losses and post-training alignments are highly recommended to compensate for this deficiency.

The positive aspect of ScaleBiO is that it helps reweight data more effectively, thus allowing the training cost of large language models to be further reduced.

\section*{Ethical Considerations}
In conducting our experiments on a diverse set of datasets for instruction following, we have given careful consideration to ethical concerns that may arise. Our work involves datasets such as ShareGPT, OpenOrca, WildChat, AlpacaChat, LMSYS-Chat, Airoboros, etc. We list the license for each dataset in the Appendix and ensure compliance with the licensing agreements for each dataset. 
Furthermore, all these data sources are publicly available and do
not involve privacy issues. 

\bibliography{custom}

\newpage
\appendix


\onecolumn

\section{Additional Experiments}

\subsection{Data Weights across Model Families}
\label{appendix:model_family_weights}

Table~\ref{tab:appendix:model-family-weight} shows the learned data weights from ScaleBiO for different backbone models under the instruction following setting.


\begin{table*}[!h]
\small
	\renewcommand    
	\arraystretch{1.3}
	\centering
            \setlength{\tabcolsep}{1.5mm}
\resizebox{\textwidth}{!}{
\begin{tabular}{lrlrlrlrlrlr}
\toprule
\multicolumn{2}{c}{Llama-3-8B} & \multicolumn{2}{c}{Llama-3-13B\tablefootnote{\url{https://huggingface.co/Replete-AI/Llama-3-13B}}} & 
\multicolumn{2}{c}{Qwen-2-7B} & 
\multicolumn{2}{c}{Gemma-2-9B} & 
\multicolumn{2}{c}{GPT-NeoX-20B} & \multicolumn{2}{c}{Yi-34B} \\ \cmidrule(r){1-2}  \cmidrule(r){3-4} \cmidrule(r){5-6} \cmidrule(r){7-8} \cmidrule(r){9-10} \cmidrule(r){11-12}
source & weight &
source & weight &
source & weight &
source & weight &
source & weight &
source & weight
\\ \midrule
WildChat        & 0.711       & WildChat         & 0.711       & 
SlimOrca         & 0.945  &
Alpaca-pt         & 0.198 &
Airoboros         & 0.986        & ShareGPT4      & 0.627     \\
Airoboros       & 0.154       & ShareGPT4        & 0.137       & 
LMSYS-Chat        & 0.008    &
Alpaca-ko         & 0.180    &
ShareGPT4         & 0.005        & 
Airoboros      & 0.111     \\
ChatAlpaca      & 0.119       & 
ChatAlpaca       & 0.021       & 
ShareGPT4       & 0.004       &
Alpaca-it        & 0.080       &
ChatAlpaca        & 0.003        & WildChat       & 0.105     \\

Total     & 0.984 &
Total     & 0.869 &
Total     & 0.957 &
Total     & 0.458 &
Total     & 0.994 &
Total     & 0.843     \\ \bottomrule
\end{tabular}
}
\caption{Data sources with \textbf{top-3} weights for LLaMA-3-8B, LLaMA-3-13B, Qwen-2-7B, Gemma-2-9B, GPT-NeoX-20B and Yi-34B in \textbf{Instruction Following} tasks.}
\label{tab:appendix:model-family-weight}
\end{table*}

\begin{table*}[!h]
\small
	\renewcommand    
	\arraystretch{1.3}
	\centering
            \setlength{\tabcolsep}{1.5mm}
\resizebox{\textwidth}{!}{
\begin{tabular}{lrlrlrlr}
\toprule
\multicolumn{2}{c}{Llama-3-8B} &  
\multicolumn{2}{c}{Qwen-2-7B} & 
\multicolumn{2}{c}{Gemma-2-9B} & 
\\ \cmidrule(r){1-2}  \cmidrule(r){3-4} \cmidrule(r){5-6}
source & weight &
source & weight &
source & weight
\\ \midrule
\texttt{TIGER-Lab/MathInstruct} & 0.131
& \texttt{TIGER-Lab/MathInstruct} & 0.132 & 
\texttt{TIGER-Lab/MathInstruct} & 0.121
\\
\texttt{teknium/GPT4-LLM-Cleaned} & 0.119 & \texttt{ise-uiuc/Magicoder-Evol-Instruct-110K} & 0.125
& DART-Math & 0.114 
\\
\texttt{anon8231489123/ShareGPT\_Vicuna\_unfiltered}      & 0.107 
& \texttt{teknium/GPTeacher-General-Instruct} & 0.102
& \texttt{openbmb/UltraInteract\_sft} & 0.110
\\
Total     &  0.357 &
Total     &  0.359 &
Total     &  0.345 &
\\ \bottomrule
\end{tabular}
}
\caption{Data sources with \textbf{top-3} weights for LLaMA-3-8B, Qwen-2-7B, Gemma-2-9B in \textbf{Mathematical Reasoning} tasks.}
\label{tab:appendix:model-family-weight_math}
\end{table*}

\subsection{Mathematical Reasoning: Stronger Benchmarks}
We conducted additional experiments on mathematical reasoning using stronger benchmarks and a smaller but higher-quality dataset.

\paragraph{Setup} Specifically, we collect 8K prompts uniformly from DART-Math \citep{tong2024dartmath}, Ultra-Interact \citep{yuan2024advancing}, MathInstruct \citep{yue2023mammoth}, and Orca-Math \citep{mitra2024orcamath}.
We then use Deepseek-R1 \citep{deepseekai2025deepseekr1incentivizingreasoningcapability} to generate responses with thinking paths to construct question-answer pairs.
After obtaining the data, we select 4K training samples using the ScaleBiO method alongside baseline methods and fine-tune the DeepSeek-R1-Distill-Qwen-1.5B model \citep{deepseekai2025deepseekr1incentivizingreasoningcapability}.
The fine-tuned models are then evaluated on the reference sets of AIME24, AIME25, and AIMO25, which contain 30, 30, and 10 questions, respectively.

\begin{table}[h!]
\centering
\resizebox{\linewidth}{!}{
\begin{tabular}{lccc|ccc}
\toprule
\multicolumn{1}{c}{} & \multicolumn{3}{c|}{\textbf{Method (pass@1)}} & \multicolumn{3}{c}{\textbf{Method (cons@64)}} \\
\textbf{Method} & \textbf{AIME 2024} & \textbf{AIME 2025} & \textbf{AIMO 2025} &
\textbf{AIME 2024} & \textbf{AIME 2025} & \textbf{AIMO 2025} \\
\midrule
Uniform   & 26.7 & 20.0 & 10.0 & \textbf{33.3} & 33.3 & \textbf{30.0} \\
LESS      & 26.7 & 20.0 & 10.0 & \textbf{33.3} & \textbf{36.7} & \textbf{30.0} \\
RHO-LOSS  & 30.0 & 20.0 & \textbf{20.0} & \textbf{33.3} & 33.3 & \textbf{30.0} \\
ScaleBiO  & \textbf{33.3} & \textbf{26.7} & \textbf{20.0} & \textbf{33.3} & \textbf{36.7} & \textbf{30.0} \\
\bottomrule
\end{tabular}}
\caption{Comparison of methods on AIME2024, AIME2025,  and AIMO2025 datasets under pass@1 and cons@64 metrics for DeepSeek-R1-Distill-Qwen-1.5B \citep{deepseekai2025deepseekr1incentivizingreasoningcapability}.}
\label{tab:combined_aimo_metrics}
\end{table}

\paragraph{Results} As shown in Table~\ref{tab:combined_aimo_metrics}, ScaleBiO consistently outperforms the baseline methods across the three benchmarks under the \textbf{pass@1} metric.
For the \textbf{Cons@64} accuracy, ScaleBiO achieves performance comparable to the baselines. We conjecture that the narrowing gap is due to the limited diversity of the small-sized dataset, which we expect to improve with the inclusion of a larger amount of data.
In summary, ScaleBiO demonstrates stable and competitive performance on challenging benchmarks across different evaluation metrics, highlighting the effectiveness of our data selection method.

\section{Experimental Details}
\subsection{Small Scale Experiments}
\label{sec:appendix:small_scale_exp_setup}

Throughout our small-scale experiments, we use GPT-2~\cite{radford2019language} with 124 million parameters as the backbone model. For bilevel optimization hyperparameters, we set the learning rate to $10^{-2}$ for sampling weights $\lambda$ and $10^{-5}$ for models $u, w$. We run our algorithm for 3 epochs with a batch size of 64 and alpha of 10 while adopting AdamW~\cite{loshchilov2017decoupled} for optimization. 

\subsection{Large Scale Experiments}
\label{sec:appendix:large_scale_exp_setup}
\begin{table}[h]
    	\centering

	\renewcommand    
	\arraystretch{1.3}
            \setlength{\tabcolsep}{1.5mm}
            
    \begin{tabular}{lrrr}
        \toprule
        Datasets & \#Samples & Kind & License \\
        \hline
        AlpacaGPT4~\citep{peng2023instruction} & 52K & Instruction & Apache-2.0 \\
        ShareGPT4~\citep{vicuna2023} & 6K & Conversation & Apache-2.0 \\
        SlimOrca~\citep{SlimOrca} & 518K & Instruction & MIT \\
        AlpacaChat~\citep{ChatAlpaca} & 20K & Conversation  & Apache-2.0 \\
        OpenOrcaGPT4~\citep{mukherjee2023orca} & 1M & Instruction & MIT \\
        WildChat~\citep{zhao2024wildchat} & 1M & Conversation & AI2 ImpACT \\
        LMSYS-Chat~\citep{zheng2023lmsyschat1m} & 1M & Conversation & LMSYS-Chat-1M \\
        GPTeacher~\citep{gpteacher} & 89K & Instruction  & MIT\\
        Airoboros~\citep{airoboros} & 59K & Conversation & CC-BY-4.0 \\
        Alpaca-es\tablefootnote{\url{https://huggingface.co/datasets/bertin-project/alpaca-spanish}} & 52K &Instruction & CC-BY-4.0 \\
        Alpaca-de\tablefootnote{\url{https://huggingface.co/datasets/mayflowergmbh/alpaca-gpt4_de}} & 50K &Instruction & Apache-2.0\\
        Alpaca-ja\tablefootnote{\url{https://huggingface.co/datasets/fujiki/japanese_alpaca_data}} & 52K &Instruction & CC-BY-NC-SA-4.0\\
        Alpaca-ko\tablefootnote{\url{https://huggingface.co/datasets/Bingsu/ko_alpaca_data}} & 50K &Instruction & CC-BY-NC-4.0\\
        Alpaca-ru\tablefootnote{\url{https://huggingface.co/datasets/IlyaGusev/ru_turbo_alpaca}} & 30K &Instruction & CC-BY-4.0 \\
        Alpaca-it\tablefootnote{\url{https://huggingface.co/datasets/mchl-labs/stambecco_data_it}} & 52K &Instruction & CC-BY-NC-SA-4.0\\
        Alpaca-fr\tablefootnote{\url{https://huggingface.co/datasets/jpacifico/French-Alpaca-dataset-Instruct-55K}} & 55K &Instruction & Apache-2.0\\\
        Alpaca-zh\tablefootnote{\url{https://huggingface.co/datasets/llm-wizard/alpaca-gpt4-data-zh}} & 49K &Instruction  & CC-BY-4.0 \\
        Alpaca-pt\tablefootnote{\url{https://huggingface.co/datasets/dominguesm/alpaca-data-pt-br}} & 52K &Instruction & CC-BY-NC-4.0
        \\ \bottomrule
    \end{tabular}
    \caption{Training data sources for the \textbf{Instruction Following} task.}
    \label{tab:appendix:data_info_inst_following}
\end{table}

\paragraph{Instruction Following}
Our training data consists of 18 distinct sources as detailed in Table~\ref{tab:appendix:data_info_inst_following}. We collect 9 high-quality datasets and 9 multilingual Alpaca datasets which serve as irrelevant data sources. For each data source, we preprocess by filtering out conversations/instructions that exceed the max length (1024 tokens in our experiments). For our reference dataset $\mD_\text{val}$ that corresponds to loss $L_1$, we prompt GPT4 using the prompt

\textit{"Help me generate 3 sets of 2-turn instructions to evaluate the \{category\} ability of LLMs. The instructions for the second turn need to be highly relevant to the first turn. The following is an example.\textbackslash n\textbackslash n\textbackslash n EXAMPLE:\{example\}\textbackslash n TURN1:\{turn1\}\textbackslash n TURN2:\{turn2\}\textbackslash n"}.

Here \textit{\{category\}} represents one of the 8 categories in MT-Bench and \textit{\{example\}} is one example from MT-Bench. In this way, we obtain a reference dataset with 1,200 samples with a similar distribution to MT-Bench. Furthermore, additional 600 samples generated in similar fashions are adopted for hyperparameter tuning for all methods.

Concerning the data reweighting and training process, we first sample 3,000 data from each source for reweighting. Then we sample 10,000 data according to the weights at the end of bilevel optimization to train the backbone model.

For ScaleBiO, the data reweighting process lasts for 3 epochs with $\alpha$ equals to 100 and initial learning rate $10^{-2}$ for weights $\lambda$. The learning rates of models $u, w$ are set to be the same and searched in range $\{ 10^{-6}, 2\times10^{-6}, 3\times 10^{-6}, 4\times10^{-6}, 5\times 10^{-6}, 6\times 10^{-6}, 8\times10^{-6}, 10^{-5} \}$. For all the fine-tuning processes, we train the LLM for 1 epoch with an initial learning rate of $8\times 10^{-6}$ and a global batch size of 64. Throughout our experiments, we adopt randomized coordinate descent with AdamW~\citep{pan2024lisa} and bfloat16 precision for efficient training and inference. Our experiments are conducted on 8 NVIDIA H100 80GB GPUs, where the total computational cost is around $\sim$6K GPU hours. The multi-GPU feature of ScaleBiO is enabled by Pytorch's FSDP~\citep{zhao2023pytorchfsdp}.

For baselines, all of them are free to utilize the additional 1,800 MT-Bench-styled samples to ensure a fair comparison with ScaleBiO, where
\begin{itemize}
  \item Uniform Weighting: directly sample 10,000 / 18 $\approx$ 5556 samples from each source, along with the additional MT-styled 1,200 samples to conduct supervised fine-tuning.
  \item LESS: we stick to settings in its original paper~\citep{xia2024less}, which adopts a warm-up training setup of learning rate $10^{-5}$, batch size 32, maximum sequence length of 1024, number of epochs 4, optimizer Adam with linear decay learning rate schedule. The LoRA setup is also similar, with $r = 128$, $\alpha = 512$, dropout = 0.1.
  \item RHO-LOSS: a training setup of learning rate $10^{-5}$, batch size 32, maximum sequence length 1024, number of epochs 1, optimizer of Adam with cosine decay learning rate schedule. Here the same reference model Qwen-2-1.5B is employed for different settings, which according to the original paper~\citep{mindermann2022rholoss}, is fine given the algorithm's non-sensitiveness to the reference model.
\end{itemize}

\paragraph{Mathematical Reasoning} 

\begin{table}[h]
    	\centering

	\renewcommand    
	\arraystretch{1.3}
            \setlength{\tabcolsep}{1.5mm}

    \resizebox{\textwidth}{!}{
    \begin{tabular}{lrrr}
        \toprule
        Datasets & \#Samples & Kind & License \\
        \hline
        DART-Math\tablefootnote{\url{https://huggingface.co/datasets/hkust-nlp/dart-math-uniform}}~\citep{tong2024dartmath} & 591K & Math & MIT \\
        SlimOrca\tablefootnote{\url{https://huggingface.co/datasets/Open-Orca/SlimOrca}}~\citep{SlimOrca} & 518K & Instruction & MIT \\
        \texttt{openbmb/UltraInteract\_sft}\tablefootnote{\url{https://huggingface.co/datasets/openbmb/UltraInteract_sft}}~\citep{yuan2024advancing} & 289K & Reasoning & MIT \\
        \texttt{TIGER-Lab/MathInstruct}\tablefootnote{\url{https://huggingface.co/datasets/TIGER-Lab/MathInstruct}}~\citep{yue2023mammoth} & 262K & Reasoning & MIT \\
        \texttt{microsoft/orca-math-word-problems-200k}\tablefootnote{\url{https://huggingface.co/datasets/microsoft/orca-math-word-problems-200k}}~\citep{mitra2024orcamath} & 200K & Math & MIT \\
        \texttt{WizardLMTeam/WizardLM\_evol\_instruct\_V2\_196k}\tablefootnote{\url{https://huggingface.co/datasets/WizardLMTeam/WizardLM_evol_instruct_V2_196k}} & 196K & Instruction & MIT \\
        \texttt{ise-uiuc/Magicoder-Evol-Instruct-110K}\tablefootnote{\url{https://huggingface.co/datasets/ise-uiuc/Magicoder-Evol-Instruct-110K}} & 110K & Coding & Apache-2.0 \\
        \texttt{anon8231489123/ShareGPT\_Vicuna\_unfiltered}\tablefootnote{\url{https://huggingface.co/datasets/anon8231489123/ShareGPT_Vicuna_unfiltered}} & 94K & Instruction & Apache-2.0 \\
        \texttt{teknium/GPTeacher-General-Instruct}\tablefootnote{\url{https://huggingface.co/datasets/teknium/GPTeacher-General-Instruct}} & 89K & Instruction & MIT
 \\
        \texttt{teknium/GPT4-LLM-Cleaned}\tablefootnote{\url{https://huggingface.co/datasets/teknium/GPT4-LLM-Cleaned}} & 55K & Instruction & Apache-2.0 \\
        \midrule
        GSM8K~\citep{cobbe2021gsm8k} & 7.5K & Math & MIT \\
        Competition Math\tablefootnote{\url{https://huggingface.co/datasets/EleutherAI/hendrycks_math}}~\citep{hendrycksmath2021} & 12.5K & Math & MIT \\
        \texttt{bigcode/self-oss-instruct-sc2-exec-filter-50k}\tablefootnote{\url{https://huggingface.co/datasets/bigcode/self-oss-instruct-sc2-exec-filter-50k}} & 50.7K & Coding & ODC-By \\
        \texttt{cais/mmlu}\tablefootnote{\url{https://huggingface.co/datasets/cais/mmlu}}~\citep{hendrycks2020mmlu,hendrycks2021ethics} & 116K & Science & MIT \\
        ARC-Easy~\citep{clark2018arc} & 5.2K & Instruction & CC-BY-SA-4.0 \\
        ARC-Challenge~\citep{clark2018arc} & 2.6K & Instruction & CC-BY-SA-4.0
        \\ \bottomrule
    \end{tabular}
    }
    \caption{Training (above the line) and validation data sources (below the line) for the \textbf{Mathematical Reasoning} task.}
    \label{tab:appendix:data_info_math_reasoning}
\end{table}
The validation set comes from the validation sets (if available) or training sets (if validation is not available) from the validation sources presented in Table~\ref{tab:appendix:data_info_math_reasoning}, where 280 samples are randomly chosen from each source and form a validation set with $280 \times 6 = 1680$ samples. For ScaleBiO, the dataset is proportionally split into two sets with $1,400$ samples and $280$ samples individually, where the former is treated as the $\mathcal{D}_\text{val}$ for $L_1$ in reweighting and the latter is adopted for hyperparameter tuning. The whole validation set is available to other baselines. Other settings and statistics remain the same as the instruction-following task.

\section{Important Lemmas}
\label{sec:appendix:lem}

Suppose Assumption~\ref{assumpt:v2} hold, the functions $\cL^{\alpha}(\lambda, w, u)$ and  $\Gamma^{\alpha}(\lambda)$ satisfy the following properties. 
\begin{lem}
\label{lem: smoothness_minmax_prob}
Under Assumption~\ref{assumpt:v2}, the followings hold:
\begin{itemize}
\item[(i)] $\cL^{\alpha}(\lambda, w, u)$ is $\mu_2\alpha$-strongly concave w.r.t. $u$; 
\item[(ii)]  $\cL^{\alpha}(\lambda, w, u)$ is $\mu_2\alpha/2$-strongly convex w.r.t. $w$ if $\alpha > 2\ell_{11}/\mu_2$.
\end{itemize}
\end{lem}
The results of Lemma~\ref{lem: smoothness_minmax_prob} can be found in~\citep{kwon2023fully} and Lemma B.1 of \cite{chen2023nearoptimal}. From Lemma B.7 in \cite{chen2023nearoptimal}, the following result holds for $\Gamma^{\alpha}(\lambda)$:
\begin{lem}\label{lem:gamma:smooth}
Under Assumption~\ref{assumpt:v2}, if $\alpha > 2\ell_{11}/\mu_2$, then $\Gamma^{\alpha}(\lambda)$ is $\ell_{\Gamma}$-smooth, where $\ell_{\Gamma} = \mathcal{O}( \kappa^3)$ is a constant that is independent on $\alpha$.
\end{lem}
Moreover, the functions $w_{\ast}^{\alpha}(\lambda)$ and $u_{\ast}(\lambda)$ satisfy the following properties.
\begin{lem}\label{lem:y:ast}
Under Assumption~\ref{assumpt:v2}, we have
\begin{align*}
 \left\| w_{\ast}^{\alpha}(\lambda) - w_{\ast}(\lambda)\right\| \leq \frac{C_0}{\alpha}
\end{align*}
where $C_0 = \ell_{10}/\mu_{2}$.
\end{lem}
The result in Lemma~\ref{lem:y:ast} follows from Lemma B.2 of \cite{chen2023nearoptimal}.
\begin{lem}\label{lem:appendix:zy}
Under Assumption~\ref{assumpt:v2}, if $\alpha > 2\ell_{11}/\mu_2$,  then we have
\begin{itemize}
\item[(i)] $u_{\ast}(\lambda)$ is $\kappa$-Lipschitz continuous;
\item[(ii)] $w_{\ast}^{\alpha}(\lambda)$ is $\ell_{u_{\ast},0}$-Lipschitz continuous where $\ell_{u_{\ast},0}=3\kappa$. 
\end{itemize}
where the condition number $\kappa=\max\left\lbrace \ell_{10}, \ell_{11}, \ell_{21}, \ell_{22} \right\rbrace / \mu_2$
\end{lem}
Claim (i) in Lemma~\ref{lem:appendix:zy} can be found in Lemma 2.2 of \citep{ghadimi2018approximation} and Claim (ii) implies from Lemma 3.2 (setting $\lambda_1=\lambda_2$) of \citep{kwon2023fully}. 

\begin{lem}
\label{lem_aspt: smoothness:yz}
Under Assumption~\ref{assumpt:v2}, if $\alpha > 2\ell_{f,1}/\mu_g$, then
 $u_{\ast}(\lambda)$ is $\ell_{\nabla u_{\ast}}$-smooth where $\ell_{\nabla_{\ast}} = \mathcal{O}\left(\frac{\kappa^2}{\mu_2}\left(\ell_{21} + 1\right) \right)$
where the condition number $\kappa=\max\left\lbrace \ell_{10}, \ell_{11}, \ell_{21}, \ell_{22} \right\rbrace / \mu_2$
\end{lem}
Following Lemma A.3 of \citep{kwon2023fully} and recalling the Lipschitz continuous property of $u_{\ast}(\lambda)$ from Lemma~\ref{lem:appendix:zy},  we have this claim is correct. 

\section{Proofs of Theorem~\ref{thm:bilevel:complexity}}
\label{sec:proof:thm}
\begin{proof}
We sample the function $\cL^{\alpha}$ by the following mini-batch approximation $\cL_{D_k}^{\alpha}$ per iteration:
\begin{equation}
    \cL_{D_k}^{\alpha} := L_1(\lambda,w; D_{\text{val}}^k) + \alpha \left( L_2( \lambda, w; D_{\text{tr}}^k) - L_2(\lambda,u; D_{\text{tr}}^k) \right)
\end{equation}
where  $D_{\text{tr}}^k$ is i.i.d. from the training dataset $D_{train}$, $D_{\text{val}}^k$ are i.i.d. from the validation dataset $D_{val}$ and independent with $D_{\text{train}}$. We use $\mathcal{F}_k$ to denote the random information before the iteration $(\lambda_k,w_k, u_k )$, that is $\mathcal{F}_k:= \sigma\left(\left\lbrace (\lambda_k,\omega_k,u_k), D_{k-1}, \cdots, D_1\right\rbrace\right)$. We use $\mathcal{C}_k = \sigma\left(\left\lbrace j_1, j_2 \cdots, j_{t-1}; r_1, r_2,\cdots, r_{t-1}\right\rbrace\right)$ to denote the random information of variables $u, w$ for the randomized block coordinates before the iteration $k$. 


 We recall the iterating formula of $\lambda$ in the stochastic version of the minimax algorithm that  $\lambda_{k+1} - \lambda_k =  - \eta_{\lambda} \nabla_{\lambda} \cL_{D_k}^{\alpha}(\lambda_k,w_k, u_k )$. At each iteration, 
 \begin{align}
      \E[\nabla \cL_{D_k}^{\alpha}(\lambda_k,\omega_k,u_k) \mid \mathcal{F}_k] = \nabla \cL^{\alpha}(\lambda_k,\omega_k,u_k).
 \end{align}
By the smoothness of $\Gamma^{\alpha}$ (see Lemma~\ref{lem:gamma:smooth}), we have
\begin{align}\label{inequ:gamma:1}
\Gamma^{\alpha}(\lambda_{k+1}) & \leq \Gamma^{\alpha}(\lambda_k) + \left\langle \nabla \Gamma^{\alpha}(\lambda_k), \lambda_{k+1} - \lambda_{k}\right\rangle + \frac{\ell_{\Gamma}}{2}\left\| \lambda_{k+1} - \lambda_k \right\|^2 \notag \\
& = \Gamma^{\alpha}(\lambda_k) - \eta_{\lambda} \left\langle \nabla \Gamma^{\alpha}(\lambda_k), \nabla_{\lambda} L_{D_k}^{\alpha}(\lambda_k,\omega_k,u_k)\right\rangle + \frac{\ell_{\Gamma}  \eta_{\lambda}^2}{2}\left\| \nabla_{\lambda} \cL_{D_k}^{\alpha}(\lambda_k,\omega_k,u_k)\right\|^2.
\end{align}
Taking conditional expectation w.r.t. $\mathcal{F}_k, \mathcal{C}_k$ on the above inequality, we have
\begin{align}
& \E[\Gamma^{\alpha}(\lambda_{k+1}) \mid \mathcal{F}_k, \mathcal{C}_k] \notag \\
& \leq  \Gamma^{\alpha}(\lambda_k) - \eta_{\lambda} \left\langle \nabla \Gamma^{\alpha}(\lambda_k), \E[\nabla_{\lambda}\cL_{D_k}^{\alpha}(\lambda_k,\omega_k,u_k) \mid \mathcal{F}_k, \mathcal{C}_k]\right\rangle  + \frac{\ell_{\Gamma}  \eta_{\lambda}^2}{2}\E\left[\left\| \nabla_{\lambda} \cL_{D_k}^{\alpha}(\lambda_k,\omega_k,u_k)\right\|^2 \mid \mathcal{F}_k, \mathcal{C}_k\right] \notag \\
  & \leq \Gamma^{\alpha}(\lambda_k) - \eta_{\lambda} \left\langle \nabla \Gamma^{\alpha}(\lambda_k), \nabla_{\lambda} \cL^{\alpha}(\lambda_k,\omega_k,u_k)\right\rangle + \frac{\ell_{\Gamma}  \eta_{\lambda}^2}{2}\E\left[\left\| \nabla_{\lambda} \cL_{D_k}^{\alpha}(\lambda_k,\omega_k,u_k)\right\|^2 \mid \mathcal{F}_k, \mathcal{C}_k\right] 
\end{align}
where the inequality follows the fact that $\cL_{D_k}^{\alpha}$ is an unbiased estimation of $\cL^{\alpha}$ and 
\begin{align}
& \E\left[\left\| \nabla_{\lambda} \cL_{D_k}^{\alpha}(\lambda_k,\omega_k,u_k)\right\|^2 \mid \mathcal{F}_k, \mathcal{C}_k\right] \notag \\
& =     \E\left[\left\| \nabla_{\lambda} \cL_{D_k}^{\alpha}(\lambda_k,\omega_k,u_k) - \nabla_{\lambda} \cL^{\alpha}(\lambda_k,\omega_k,u_k) + \nabla_{\lambda} \cL^{\alpha}(\lambda_k,\omega_k,u_k) \right\|^2\mid \mathcal{F}_k \right] \notag \\
& \leq  \E\left[\left\| \nabla_{\lambda} \cL_{D_k}^{\alpha}(\lambda_k,\omega_k,u_k) - \nabla_{\lambda} \cL^{\alpha}(\lambda_k,\omega_k,u_k) \right\|^2 \mid \mathcal{F}_k \right] + \left\|\nabla_{\lambda} \cL^{\alpha}(\lambda_k,\omega_k,u_k) \right\|^2 \notag \\
& \leq \frac{\sigma_1^2 + 2\alpha^2\sigma_2^2}{B} + \left\|\nabla_{\lambda} \cL^{\alpha}(\lambda_k,\omega_k,u_k) \right\|^2
\end{align}
where the variance of the minibatch stochastic gradients (with batch size $B$) is bounded
\begin{align}
\E\left[\left\| \nabla L_1(\lambda,w; D_{\text{val}}^k) - \nabla L_1(\lambda, w) \right\|^2\right] \leq \frac{\sigma_1^2}{B}, \quad \E\left[\left\| \nabla L_2( \lambda, w; D_{\text{tr}}^k) - \nabla L_2(\lambda, w) \right\|^2\right] \leq \frac{\sigma_2^2}{B},
\end{align}
then
\begin{align}
  & \E\left[\left\| \nabla_{\lambda} \cL_{D_k}^{\alpha}(\lambda_k,\omega_k,u_k) - \nabla_{\lambda} \cL^{\alpha}(\lambda_k,\omega_k,u_k) \right\|^2 \mid \mathcal{F}_k \right] \notag \\
  & =   \E\left[\left\| \nabla_{\lambda} L_1(\lambda_k, w_k; D_{\text{val}}^k) - \nabla_{\lambda} L_1(\lambda_k, w_k) \right\|^2 +  \alpha^2 \left\| \nabla_{\lambda} L_2(\lambda_k, w_k; D_{\text{tr}}^k) - \nabla_{\lambda} L_2( \lambda_k, w_k) \right\|^2\mid \mathcal{F}_k \right]  \notag \\
  & \quad + \alpha^2 \E\left[ \left\| \nabla_{\lambda} L_2( \lambda_k, u_k; D_{\text{tr}}^k) - \nabla_{\lambda} L_2(\lambda_k, u_k) \right\|^2\mid \mathcal{F}_k \right] \notag \\
  & \leq \frac{\sigma_1^2 + 2\alpha^2\sigma_2^2}{B}.
\end{align}
Applying the above results, we have
\begin{align}\label{inequ:main:expect}
\E[\Gamma^{\alpha}(\lambda_{k+1}) \mid \mathcal{F}_k, \mathcal{C}_k] 
& \leq \Gamma^{\alpha}(\lambda_k) -  \eta_{\lambda} \left\langle \nabla \Gamma^{\alpha}(\lambda_k), \nabla_{\lambda} \cL^{\alpha}(\lambda_k,\omega_k,u_k)\right\rangle  + \frac{\ell_{\Gamma}  \eta_{\lambda}^2}{2}\left\|\nabla_{\lambda} \cL^{\alpha}(\lambda_k,\omega_k,u_k) \right\|^2 \notag \\
& + \frac{\ell_{\Gamma}  \eta_{\lambda}^2}{2B}\left(\sigma_1^2 + 2\alpha^2\sigma_2^2\right).
\end{align}
Let $\delta_k = \left\|u_k - u_{\ast}(\lambda_k) \right\|^2$ and $r_k = \left\| w_k - w_{\ast}^{\alpha}(\lambda_k) \right\|^2$.  The inner product term of RHS of \eqref{inequ:main:expect} is estimated as follows:
\begin{align}\label{inequ:inner:prod}
& - \left\langle \nabla \Gamma^{\alpha}(\lambda_k), \nabla_{\lambda} \cL^{\alpha}(\lambda_k,\omega_k,u_k)\right\rangle \notag \\
 = &  - \left\langle \nabla \Gamma^{\alpha}(\lambda_k), \nabla_{\lambda} \cL^{\alpha}(\lambda_k,\omega_k,u_k) - \nabla_{\lambda} \cL^{\alpha}(\lambda_k, w_k, u_{\ast}(\lambda_k)) \right\rangle \notag \\
& - \left\langle \nabla \Gamma^{\alpha}(\lambda_k), \nabla_{\lambda}\cL^{\alpha}(\lambda_k, w_k, u_{\ast}(\lambda_k)) - \nabla_{\lambda} \Phi^{\alpha}(w_{\ast}^{\alpha}(\lambda_k), \lambda_k) + \nabla_{\lambda} \Phi^{\alpha}(w_{\ast}^{\alpha}(\lambda_k), \lambda_k)\right\rangle  \notag \\
\mathop{=}^{(a)} & - \left\langle \nabla \Gamma^{\alpha}(\lambda_k), \nabla_{\lambda} \cL^{\alpha}( \lambda_k, w_k, u_k) - \nabla_{\lambda} \cL^{\alpha}(\lambda_k, w_k, u_{\ast}(\lambda_k)) \right\rangle \notag \\
& - \left\langle \nabla \Gamma^{\alpha}(\lambda_k), \nabla_{\lambda} \Phi^{\alpha}(\lambda_k, w_k) - \nabla_{\lambda} \Phi^{\alpha}(\lambda_k, w_{\ast}^{\alpha}(\lambda_k)) + \nabla \Gamma^{\alpha}(\lambda_k)\right\rangle  \notag \\
& = - \left\| \nabla \Gamma^{\alpha}(\lambda_k) \right\|^2 -\left\langle \nabla \Gamma^{\alpha}(\lambda_k), \nabla_{\lambda} \cL^{\alpha}(u_k, \omega_k, \lambda_k) - \nabla_{\lambda} \cL^{\alpha}(\lambda_k, w_k, u_{\ast}(\lambda_k)) \right\rangle  \notag \\
& - \left\langle \nabla \Gamma^{\alpha}(\lambda_k), \nabla_{\lambda} \cL^{\alpha}(\lambda_k, w_k, u_{\ast}(\lambda_k)) - \nabla_{\lambda} \cL^{\alpha}(\lambda_k, w_{\ast}^{\alpha}(\lambda_k), u_{\ast}(\lambda_k))\right\rangle \notag \\
\mathop{\leq}^{(b)}&  - \frac{1}{2}\left\| \nabla \Gamma^{\alpha}(\lambda_k) \right\|^2 + \left\|\nabla_{\lambda}\cL^{\alpha}(\lambda_k, w_k, u_k) - \nabla_{\lambda} \cL^{\alpha}(\lambda_k, w_k, u_{\ast}(\lambda_k)  ) \right\|^2 \notag \\
& \quad + \left\|  \nabla_{\lambda} \cL^{\alpha}(\lambda_k, w_k, u_{\ast}(\lambda_k)) - \nabla_{\lambda} \cL^{\alpha}(\lambda_k, w_{\ast}^{\alpha}(\lambda_k), u_{\ast}(\lambda_k))\right\|^2 \notag \\
\mathop{\leq}^{(c)} &  - \frac{1}{2}\left\| \nabla \Gamma^{\alpha}(\lambda_k) \right\|^2 + \alpha^2\ell_{21}^2 \left\|u_k - u_{\ast}(\lambda_k) \right\|^2 + 2\left(\ell_{11}^2 +  \alpha^2\ell_{21}^2 \right) \left\| \omega_k - \omega_{\ast}^{\alpha}(\lambda_k) \right\|^2 \notag \\
=& - \frac{1}{2}\left\| \nabla \Gamma^{\alpha}(\lambda_k) \right\|^2 + \alpha^2\ell_{21}^2 \delta_k +  2\left(\ell_{11}^2 +  \alpha^2\ell_{21}^2 \right) r_k
\end{align}
where $(a)$ uses the optimality of $\Phi$ over $w$ that $ \nabla_{\lambda} \Phi^{\alpha}(w_{\ast}^{\alpha}(\lambda_k), \lambda_k) = \nabla \Gamma^{\alpha}(\lambda_k) = \nabla_{\lambda} \cL^{\alpha}(\lambda_k, w_{\ast}^{\alpha}(\lambda_k), u_{\ast}(\lambda_k))$, $(b)$ follows from the Cauchy-Schwartz inequality and $(c)$ uses the smoothness of $L_1$ and $L_2$. 
Next we turn to estimate the norm of gradient $\nabla_{\lambda} \cL^{\alpha}(\lambda_k, w_k, u_k)$ as follows
\begin{align}\label{inequ:norm:grad}
\left\| \nabla_{\lambda} \cL^{\alpha}(\lambda_k, w_k, u_k)\right\|^2  & = \left\| \nabla_{\lambda} \cL^{\alpha}(\lambda_k, w_k, u_k) - \nabla \Gamma^{\alpha}(\lambda_k) + \nabla \Gamma^{\alpha}(\lambda_k) \right\|^2 \notag \\ & \leq 2 \left(\left\| \nabla \Gamma^{\alpha}(\lambda_k) \right\|^2  + \left\|\nabla_{\lambda} \cL^{\alpha}(\lambda_k, w_k, u_k) - \nabla \Gamma^{\alpha}(\lambda_k) \right\|^2 \right) \notag \\
& \leq 2\left\| \nabla \Gamma^{\alpha}(\lambda_k) \right\|^2  + 4\left\|\nabla_{\lambda} \cL^{\alpha}(\lambda_k, w_k, u_k) - \nabla_{\lambda} \cL^{\alpha}(\lambda_k, w_k, u_{\ast}(\lambda_k))\right\|^2 \notag \\
& \quad + 4\left\|  \nabla_{\lambda} \cL^{\alpha}(\lambda_k, w_k, u_{\ast}(\lambda_k)) - \nabla_{\lambda} \cL^{\alpha}(\lambda_k, w_{\ast}^{\alpha}(\lambda_k), u_{\ast}(\lambda_k)) \right\|^2 \notag \\
& \mathop{\leq}^{(a)} 2 \left\| \nabla \Gamma^{\alpha}(\lambda_k) \right\|^2 + 4\alpha^2\ell_{11}^2 \left\|u_k - u_{\ast}(\lambda_k) \right\|^2 + 8\left(\ell_{11}^2 +  \alpha^2\ell_{21}^2 \right) \left\| \omega_k - \omega_{\ast}^{\alpha}(\lambda_k) \right\|^2 \notag \\
& = 2\left\| \nabla \Gamma(\lambda_k) \right\|^2 + 4\alpha^2\ell_{11}^2 \delta_k + 8\left(\ell_{11}^2 +  \alpha^2\ell_{21}^2 \right) r_k
\end{align}
where $(a)$ uses the smoothness of objectives $L_1, L_2$.
 Incorporating the above inequalities (\ref{inequ:inner:prod}) and (\ref{inequ:norm:grad}) into \eqref{inequ:main:expect} gives
\begin{align}\label{inequ:gammak}
\E[\Gamma^{\alpha}(\lambda_{k+1}) \mid \mathcal{F}_k, \mathcal{C}_k ] 
& \leq \Gamma^{\alpha}(\lambda_k) - \frac{\eta_{\lambda}}{2}\left\| \nabla \Gamma^{\alpha}(\lambda_k) \right\|^2  + \frac{\ell_{\Gamma}  \eta_{\lambda}^2}{2}\left(2\left\| \nabla \Gamma^{\alpha}(\lambda_k) \right\|^2 + 4\alpha^2\ell_{11}^2 \delta_k + 8\left(\ell_{11}^2 +  \alpha^2\ell_{21}^2 \right) r_k\right) \notag \\
&  + \eta_{\lambda} \left(\alpha^2\ell_{11}^2 \delta_k +  2\left(\ell_{11}^2 +  \alpha^2\ell_{21}^2 \right) r_k\right) + \frac{\ell_{\Gamma}  \eta_{\lambda}^2}{2}\left(\sigma_1^2 + 2\alpha^2\sigma_2^2\right).
\end{align}
Then, we focus on estimating $\delta_k$ and $r_k$. For the inner variables $u, w$, we use the randomized block coordinates method with total $J$ blocks and each block is uniformly chosen. By the strong concavity of $\cL^{\alpha}$ with respect to $u$, we first achieve the following evaluations for $\delta_k$:
\begin{align}\label{inequ:uk}
 & \E\left[\left\|  u_{k+1} - u_{\ast}( \lambda_{k}) \right\|^2 \mid \mathcal{F}_k, \mathcal{C}_k \right] = \E\left[\left\|u_k - \alpha\eta_{u} U_{j_t}\nabla_{u} L_2\left(\lambda_k, u_k; \mathcal{D}_{k}^{\text{tr}}\right) - u_{\ast}(\lambda_{k})  \right\|^2  \mid \mathcal{F}_k, \mathcal{C}_k\right]\notag \\
  & = \left\|u_{\ast}( \lambda_{k}) - u_k \right\|^2 - 2\alpha\eta_{u}\E\left[\left\langle u_k- u_{\ast}( \lambda_{k}), \nabla_{u} L_2(\lambda_k, u_k; \mathcal{D}_{k}^{\text{tr}})\right\rangle_{j_t} \mid\mathcal{F}_k, \mathcal{C}_k\right] \notag \\
  & \quad + \alpha^2\eta_{u}^2 \E\left[\left\| U_{j_t}\nabla_{u} L_2(u_k, \lambda_k; \mathcal{D}_{k}^{\text{tr}}) \right\|^2 \mid \mathcal{F}_k, \mathcal{C}_k\right] \notag \\
    & \mathop{=}^{(a)} \left\|u_{\ast}( \lambda_{k}) - u_k \right\|^2 - \frac{2\alpha\eta_{u}}{J}\left\langle u_k- u_{\ast}( \lambda_{k}),  \nabla_{u} L_2( \lambda_k, u_k)\right\rangle + \frac{\alpha^2\eta_{u}^2}{J} \E\left[\left\| \nabla_{u} L_2(\lambda_k, u_k; \mathcal{D}_{k}^{\text{tr}})\right\|^2 \mid \mathcal{F}_k \right] \notag \\
  & \mathop{\leq}^{(b)} \left\|u_{\ast}( \lambda_{k}) - u_k \right\|^2 - \frac{2\eta_u\alpha}{J}\left( L_2(\lambda_k, u_k)-L_2(\lambda_k, u_{\ast}(\lambda_k)) + \frac{\mu_2}{2}\left\|u_{\ast}(\lambda_k) - u_k \right\|^2 \right) \notag \\ 
  & \quad +  \frac{\alpha^2\eta_{u}^2}{J} \E\left[\left\| \nabla_{u} L_2(\lambda_k, u_k; \mathcal{D}_{k}^{\text{tr}})\right\|^2 \mid \mathcal{F}_k \right]  \notag \\
  & \mathop{=}^{(c)}  \left\|u_{\ast}( \lambda_{k}) - u_k \right\|^2 - \frac{2\eta_u\alpha}{J}\left(L_2(\lambda_k, u_k)-L_2(\lambda_k, u_{\ast}(\lambda_k))+ \frac{\mu_2}{2}\left\|u_{\ast}(\lambda_k) - u_k \right\|^2 \right) \notag \\ 
  & \quad +  \frac{\alpha^2\eta_{u}^2}{J} \E\left[\left\| \nabla_{u} L_2(\lambda_k, u_k; \mathcal{D}_{k}^{\text{tr}}) - \nabla_{u} L_2(\lambda_k, u_k)\right\|^2 \mid \mathcal{F}_k \right] +  \frac{\alpha^2\eta_{u}^2}{J}\left\| \nabla_{u} L_2(\lambda_k, u_k)\right\|^2  \notag \\
& \mathop{\leq}^{(d)}  \left\|u_{\ast}( \lambda_{k}) - u_k \right\|^2 - \frac{2\eta_u\alpha}{J}\left(L_2(\lambda_k, u_k)-L_2(\lambda_k, u_{\ast}(\lambda_k))  + \frac{\mu_2}{2}\left\|u_{\ast}(\lambda_k) - u_k \right\|^2 \right) \notag \\ 
  & \quad +  \frac{\alpha^2\eta_{u}^2}{J} \E\left[\left\| \nabla_{u} L_2(\lambda_k, u_k; \mathcal{D}_{k}^{\text{tr}}) - \nabla_{u} L_2(\lambda_k, u_k)\right\|^2 \mid \mathcal{F}_k \right]+  \frac{2\ell_{21} \eta_u^2\alpha^2}{J}\left(L_2(\lambda_k, u_k) -  L_2(\lambda_k, u_{\ast}(\lambda_k))\right) \notag \\
  & \mathop{\leq}^{(e)} \left(1-\frac{\alpha\mu_2 \eta_u}{J} \right) \left\|u_{\ast}(\lambda_k) - u_k \right\|^2 + \frac{\alpha^2\eta_u^2\sigma_2^2}{J B} .
\end{align}
where $(a)$ use the truth that since the $j_k$ block coordinate is uniformly chosen from $\left\lbrace 1, 2,\cdots, J \right\rbrace$, we have 
\begin{align}
\E\left[\left\langle u_k-u_{\ast}( \lambda_{k}) , \nabla_{u} L_2(\lambda_k, u_k; \mathcal{D}_{k}^{\text{tr}})\right\rangle_{j_t} \mid\mathcal{F}_k, \mathcal{C}_k\right] &  = \frac{1}{J}\E\left[\left\langle u_k-u_{\ast}( \lambda_{k}),  \nabla_{u} L_2( \lambda_k, u_k; \mathcal{D}_{k}^{\text{tr}})\right\rangle \mid\mathcal{F}_k\right] \notag \\
& = \frac{1}{J}\left\langle u_k-u_{\ast}( \lambda_{k}),  \nabla_{u} L_2( \lambda_k, u_k)\right\rangle
\end{align}
and 
\begin{align}
\E\left[\left\| U_{j_t}\nabla_{u} L_2(u_k, \lambda_k; \mathcal{D}_{k}^{\text{tr}}) \right\|^2 \mid \mathcal{F}_k, \mathcal{C}_k\right] = \frac{1}{J}\E\left[\left\| \nabla_{u} L_2(\lambda_k, u_k; \mathcal{D}_{k}^{\text{tr}})\right\|^2 \mid\mathcal{F}_k\right]  
\end{align}
$(b)$ follows from the strong convexity of $L_2$ w.r.t. $u$ which implies that 
\begin{align}
  L_2(\lambda_k, u_{\ast}(\lambda_k) ) \geq L_2(\lambda_k, u_k) +\left\langle \nabla_{u}L_2(\lambda_k, u_k ), u_{\ast}(\lambda_k) - u_k \right\rangle + \frac{\mu_2}{2}\left\| u_k -u_{\ast}(\lambda_k) \right\|^2  \notag ,
\end{align}
$(c)$ uses the relationship $\E\left[\nabla_u L_2(\lambda_k, u_k; D_{\text{tr}}^k)\mid \mathcal{F}_k\right] = \nabla_u L_2(\lambda_k, u_k)$ which induces that \begin{align}
   \E\left[\left\|\nabla_u L_2(\lambda_k, u_k; D_{k}^{\text{str}}) \right\|^2  \mid \mathcal{F}_k\right] =   \E\left[\left\|\nabla_u L_2(\lambda_k, u_k; \mathcal{D}_{k}^{\text{str}}) - \nabla_u L_2(\lambda_k, u_k) \right\|^2 \mid \mathcal{F}_k\right] + \left\| \nabla_u L_2(\lambda_k, u_k) \right\|^2
\end{align}
and $(d)$ uses the optimality of $u_{\ast}(\lambda)$ and the smoothness of $L_2$ such that 
\begin{align}
L_2(\lambda_k, u_{\ast}(\lambda_k)) - L_2(\lambda_k, u_k)  & \leq L_2(\lambda_k, \tilde{u}) -L_2(\lambda_k, u_k ) \notag \\
 & \leq  L_2(\lambda_k, u_k )  + \left\langle \nabla_u L_2(\lambda_k, u_k ), \tilde{u} - u_k \right\rangle + \frac{\ell_{21}}{2}\left\|\tilde{u} - u_k \right\|^2 -L_2(\lambda_k, u_k ) \notag \\
 & = - \frac{1}{2\ell_{21}} \left\|\nabla_u L_2(\lambda_k, u_k )\right\|^2
\end{align}
where $\tilde{u} = u_k - \frac{1}{\ell_{21}} \nabla_u L_2(\lambda_k, u_k )$
and $(e)$ uses 
\begin{align}
& \E\left[\left\| \nabla_{u} L_2(\lambda_k, u_k; \mathcal{D}_{k}^{\text{str}}) - \nabla_{u} L_2(\lambda_k, u_k)\right\|^2 \mid \mathcal{F}_k \right]  \leq \frac{\sigma_2^2}{B}.
\end{align}
and $\eta_u \leq 1/(\alpha \ell_{21})$.
Then we make the following recursive estimation for $\delta_k$:
\begin{align}\label{inequ:deltak:est}
\delta_{k+1} =& \left\|u_{\ast}(\lambda_{k+1}) - u_{k+1}  \right\|^2 =   \left\|u_{\ast}(\lambda_{k+1}) -  u_{\ast}(\lambda_{k}) + u_{\ast}(\lambda_{k}) - u_{k+1}  \right\|^2\notag \\ \mathop{\leq}^{(a)} & (1 + \gamma_1 ) \left\|u_{\ast}(\lambda_{k+1}) -u_{\ast}(\lambda_{k})  \right\|^2 + (1+1/\gamma_1) \left\|  u_{\ast}(\lambda_{k}) - u_{k+1}  \right\|^2 \notag \\
\mathop{\leq}^{(b)} &  (1 + \gamma_1) \kappa^2\left\| \lambda_{k+1} - \lambda_k \right\|^2 + (1+1/\gamma_1) \left\|  u_{\ast}(\lambda_{k}) - u_{k+1}  \right\|^2 \notag \\
\mathop{\leq}^{(c)} &  (1 + \gamma_1) \kappa^2\left\| \lambda_{k+1} - \lambda_k \right\|^2 + (1+1/\gamma_1)\left(\left(1-\frac{\alpha\mu_2 \eta_u}{J} \right) \delta_k +\frac{\alpha^2\eta_u^2\sigma_2^2}{J B}  \right)\notag \\
\mathop{\leq}^{(d)}&  (1 + \gamma_1) \kappa^2\eta_{\lambda}^2\left\| \nabla_{\lambda} \cL^{\alpha}(u_k, \omega_k, \lambda_k)\right\|^2 + (1+1/\gamma_1)\left(1-\frac{\alpha\mu_2 \eta_u}{J}\right) \delta_k  + (1+1/\gamma_1) \frac{\alpha^2\eta_u^2\sigma_2^2}{J B}  \notag \\
\mathop{\leq}^{(e)}&   (1 + \gamma_1) \kappa^2\eta_{\lambda}^2 \left( 2\left\| \nabla \Gamma^{\alpha}(\lambda_k) \right\|^2 + 4\alpha^2\ell_{21}^2 \delta_k + 8\left(\ell_{11}^2 +  \alpha^2\ell_{21}^2 \right) r_k \right) + (1+1/\gamma_1)\left(1-\frac{\alpha\mu_2 \eta_u}{J}\right) \delta_k \notag \\
& + (1+1/\gamma_1)\frac{\alpha^2\eta_u^2\sigma_2^2}{J B} \notag \\
=&   \left( 4\alpha^2(1 + \gamma_1) \kappa^2\eta_{\lambda}^2\ell_{21}^2 +(1+1/\gamma_1)\left(1-\frac{\alpha\mu_2 \eta_u}{J}\right)\right) \delta_k + 8(1 + \gamma_1) \kappa^2\eta_{\lambda}^2\left(\ell_{11}^2 +  \alpha^2\ell_{21}^2 \right) r_k \notag \\
\quad&  + 2(1 + \gamma_1) \kappa^2\eta_{\lambda}^2\left\| \nabla \Gamma^{\alpha}(\lambda_k) \right\|^2 + (1+1/\gamma_1)\frac{\alpha^2\eta_u^2\sigma_2^2}{J B}
\end{align}
where $(a)$ follows from Cauchy-Schwartz inequality with $\gamma_1 >0$; $(b)$ uses the Lipschitz continuity of $u_{\ast}$ from Lemma~\ref{lem:appendix:zy}; $(c)$ follows from the inequality (\ref{inequ:uk}); $(d)$ uses the iterating formula of $\lambda_{k+1}$; $(e)$ follows from the inequality (\ref{inequ:norm:grad}).

Since $L_1 + \alpha L_2$ is strongly convex with respect to $w$ with parameter $\alpha\mu_2/2$ if $\alpha \geq 2\ell_{21}/\mu_2$. Similar to $\delta_k$, we can achieve the following result for $r_k$
\begin{align}
    \E\left[\left\|\omega_{\ast}^{\alpha}(\lambda_k) - \omega_{k+1} \right\|^2 \mid \mathcal{F}_k, \mathcal{C}_k\right] \leq \left( 1- \frac{\alpha\mu_2\eta_{w}}{2J}\right)r_k  + \frac{\eta_{w}^2\left(\sigma_1^2 + \alpha^2 \sigma_2^2\right)}{JB}.
\end{align}
Following the same procedure as in~\eqref{inequ:deltak:est}, we estimate the recursion $r_k$ as below
\begin{align}\label{inequ:rk}
 r_{k+1} & \leq (1+\gamma_2)\left\| \omega_{\ast}^{\alpha}(\lambda_{k+1}) - \omega_{\ast}^{\alpha}(\lambda_k)\right\|^2 + (1+ \gamma_2^{-1}) \left\| \omega_{\ast}^{\alpha}(\lambda_k) - \omega_{k+1} \right\|^2 \notag \\
 & \leq (1+\gamma_2)\kappa^2\left\|\lambda_{k+1} - \lambda_k\right\|^2 + (1+ \gamma_2^{-1}) \left(\left( 1- \frac{\alpha\mu_2\eta_{w}}{2J}\right)r_k  + \frac{\eta_{w}^2\left(\sigma_1^2 + \alpha^2 \sigma_2^2\right)}{JB}\right)\notag \\
 & \leq \left( 4(1 + \gamma_2) \kappa^2\eta_{\lambda}^2\left(\ell_{11}^2 +  \alpha^2\ell_{21}^2 \right) +(1+1/\gamma_2)\left(1-\frac{\alpha\mu_2 \eta_w}{2J}\right)\right) r_k + 8(1 + \gamma_2) \kappa^2\eta_{\lambda}^2  \alpha^2\ell_{21}^2\delta_k \notag \\
\quad&  + 2(1 + \gamma_2) \kappa^2\eta_{\lambda}^2\left\| \nabla \Gamma^{\alpha}(\lambda_k) \right\|^2 + (1+1/\gamma_2)\frac{\eta_{w}^2\left(\sigma_1^2 + \alpha^2 \sigma_2^2\right)}{JB}
\end{align}
where $\gamma_2 > 0$. 

We define the Lyapunov function 
\begin{align}
R_k = \Gamma^{\alpha}(\lambda_{k}) - \Gamma_{\min}^{\alpha} + \xi_k^1  \delta_{k} + \xi_{2}^k  r_{k}
\end{align}
where $\xi_{k}^1, \xi_{k}^2 >0$  are non-increasing sequences and $\Gamma_{\min}^{\alpha}$ is the minimum of $\Gamma^{\alpha}$. We must have $R_k \geq 0$. Incorporating the results of \eqref{inequ:gammak}, \eqref{inequ:deltak:est}, \eqref{inequ:rk} gives
\begin{align}\label{inequ:main:lyapu}
 & \E[R_{k+1} \mid \mathcal{F}_k, \mathcal{C}_k]  \notag \\
 & \leq R_k - \frac{\eta_{\lambda}}{2}\left\| \nabla \Gamma(\lambda_k) \right\|^2  + \frac{\ell_{\Gamma}  \eta_{\lambda}^2}{2}\left(2\left\| \nabla \Gamma^{\alpha}(\lambda_k) \right\|^2 + 4\alpha^2\ell_{21}^2 \delta_k + 8\left(\ell_{11}^2 +  \alpha^2\ell_{21}^2 \right) r_k\right) \notag \\
&  + \eta_{\lambda} \left(\alpha^2\ell_{21}^2 \delta_k +  2\left(\ell_{11}^2 +  \alpha^2\ell_{21}^2 \right) r_k\right) + \frac{\ell_{\Gamma}  \eta_{\lambda}^2}{2}\left(\sigma_1^2 + 2\alpha^2\sigma_2^2\right) 
+ \left(\xi_{k+1}^1 \delta_{k+1} - \xi_k^1 \delta_k \right) + \left(\xi_{k+1}^2 r_{k+1} - \xi_k^2 r_k \right) \notag \\
& \leq R_k - \left(\frac{\eta_{\lambda}}{2} - \ell_{\Gamma}  \eta_{\lambda}^2 
 - 2\xi_{k+1}^1 (1+\gamma_1)\kappa^2\eta_{\lambda}^2 - 2\xi_{k+1}^2 (1+\gamma_2)\kappa^2\eta_{\lambda}^2\right) \left\| \nabla \Gamma^{\alpha}(\lambda_k) \right\|^2  + \phi_1 \delta_k + \phi_2 r_k \notag \\
 & + \frac{\ell_{\Gamma}  \eta_{\lambda}^2}{2}\left(\sigma_1^2 + 2\alpha^2\sigma_2^2\right)  + \xi_{k+1}^1(1+\gamma_1^{-1})\frac{\alpha^2 \eta_{u}^2\sigma_2^2}{JB} + \xi_{k+1}^2(1+\gamma_2^{-1})\frac{\eta_{w}^2\left(\sigma_1^2 + \alpha^2 \sigma_2^2\right)}{JB}
\end{align}
 where 
 \begin{align}
     \phi_1 &= \xi_{k+1}^1 \left( 4\alpha^2(1 + \gamma_1) \kappa^2\eta_{\lambda}^2\ell_{21}^2 +(1+1/\gamma_1)\left(1-\frac{\alpha\mu_2 \eta_u}{J}\right)\right) - \xi_k^1 + 2\ell_{\Gamma}\eta_{\lambda}^2\alpha^2\ell_{21}^2 + \eta_{\lambda}\alpha^2\ell_{21}^2 \notag \\
     & + 8\xi_{k+1}^2(1 + \gamma_2) \kappa^2\eta_{\lambda}^2\left(\ell_{11}^2 +  \alpha^2\ell_{21}^2 \right)  \notag \\
     \phi_2 & = \xi_{k+1}^2 \left( 4(1 + \gamma_2) \kappa^2\eta_{\lambda}^2\left(\ell_{11}^2 +  \alpha^2\ell_{21}^2 \right) +(1+1/\gamma_2)\left(1-\frac{\alpha\mu_2 \eta_w}{2J}\right)\right) - \xi_k^2 + 4\ell_{\Gamma}\eta_{\lambda}^2\left(\ell_{11}^2 +  \alpha^2\ell_{21}^2 \right) \notag \\
     & + 2\eta_{\lambda}\left(\ell_{11}^2 +  \alpha^2\ell_{21}^2 \right) + 8\xi_{k+1}^1(1 + \gamma_1) \kappa^2\eta_{\lambda}^2  \alpha^2\ell_{21}^2.
 \end{align}
 Let $\eta_u = \eta_{\omega} = \eta_0/K^a$ and $\eta_{\lambda} = \eta_{\lambda}^0/K^b$, and $\alpha = K^{c}$ where $0\leq  a \leq b$ and $c > 0$, and $\ell =\max \left\lbrace\ell_{11}, \ell_{21}\right\rbrace$ then $\phi_1$ and $\phi_2$ can be re-written as:
 \begin{align}
       \phi_1 &= \xi_{k+1}^1 \left( \frac{4(1 + \gamma_1) \kappa^2 (\eta_{\lambda}^0)^2}{K^{2(b-c)}}\ell^2 +(1+1/\gamma_1)\left(1- \frac{\mu_2\eta_0}{J K^{(a-c)}}\right)\right) - \xi_k^1 + \frac{2\ell_{\Gamma}\ell^2(\eta_{\lambda}^0)^2}{K^{2(b-c)}} + \frac{\ell^2\eta_{\lambda}^0}{K^{(b-2c)}} \notag \\
       & \quad + \frac{8\xi_{k+1}^2(1+\gamma_2)\kappa^2\ell^2(\eta_{\lambda}^0)^2}{K^{2(b-c)}} \notag \\
     \phi_2 & = 
     \xi_{k+1}^2 \left( \frac{4(1 + \gamma_2) \kappa^2 (\eta_{\lambda}^0)^2}{K^{2(b-c)}}\ell^2 +(1+1/\gamma_2)\left(1- \frac{\mu_2\eta_0}{2JK^{(a-c)}}\right)\right) - \xi_k^2 + \frac{4\ell_{\Gamma}\ell^2(\eta_{\lambda}^0)^2}{K^{2{(b-c)}}} + \frac{2\ell^2\eta_{\lambda}^0}{K^{(b-2c)}} \notag \\
     & \quad + \frac{8\xi_{k+1}^1(1+\gamma_1)\kappa^2\ell^2(\eta_{\lambda}^0)^2}{K^{2{(b-c)}}}.
 \end{align}
 In order to achieve $\phi_1 \leq 0$ and $\phi_2 \leq 0$, we might let  $\gamma_1 =\gamma_2 = 4J K^{(a-c)}/(\mu_2 \eta_0)-1$, then
 \begin{align}
   (1+1/\gamma_1)\left(1- \frac{\mu_2\eta_0}{J K^{(a-c)}}\right) &\leq   1- \frac{3\mu_2\eta_0}{4J K^{(a-c)}} \notag \\
    (1+1/\gamma_2)\left(1- \frac{\mu_2\eta_0}{2J K^{(a-c)}}\right) &\leq   1- \frac{\mu_2\eta_0}{4 J K^{(a-c)}}. 
 \end{align}
For $\eta_0 \leq \frac{8J}{\mu_2}$, we have $\frac{\mu_2\eta_0}{4J} \leq \frac{1}{2}$. Consider that  $\xi_k^1$ and $\xi_k^2$ are non-increasing sequence, then $\xi_k^1 \geq \xi_{k+1}^1$ and $\xi_k^2 \geq \xi_{k+1}^2$, we have 
 \begin{align}
        \phi_1 & \leq \xi_k^1 \left(1 + \frac{(\eta_{\lambda}^0)^2\ell^2\kappa^2 J}{\mu_2 \eta_0 K^{(2b-c-a)}}- \frac{3\mu_2\eta_0}{4 J K^{(a-c)}}\right) - \xi_k^1 + \frac{2\ell_{\Gamma}\ell^2(\eta_{\lambda}^0)^2}{K^{2(b-c)}} + \frac{\ell^2\eta_{\lambda}^0}{K^{(b-2c)}} + \xi_k^2\frac{8J(\eta_{\lambda}^0)^2\ell^2\kappa^2}{\mu_2 \eta_0 K^{(2b-c-a)}} \leq 0 \notag \\
     \phi_2 & \leq  
     \xi_k^2 \left(1+\frac{(\eta_{\lambda}^0)^2\ell^2\kappa^2J}{\mu_2\eta_0K^{(2b-c-a)}} - \frac{\mu_2\eta_0}{4 J K^{(a-c)}}\right) - \xi_k^2 + \frac{4\ell_{\Gamma}\ell^2(\eta_{\lambda}^0)^2}{K^{2(b-c)}} + \frac{2\ell^2\eta_{\lambda}^0}{K^{(b-2c)}} + \xi_k^1\frac{8J (\eta_{\lambda}^0)^2\ell^2 \kappa^2}{\mu_2\eta_0 K^{(2b-c-a)}} \leq 0 \notag
 \end{align}
 If $\eta_{\lambda}^0 \leq 1/(2\ell_{\Gamma})$ and $\eta_0/\eta_{\lambda}^0 \geq 6\sqrt{2}\kappa^2 J$, for $b \geq a$ and $k > 1$, then
 \begin{align*}
 \frac{2\ell_{\Gamma}\ell^2(\eta_{\lambda}^0)^2}{K^{2(b-c)}} \leq \frac{\ell^2\eta_{\lambda}^0}{K^{(b-2c)}},  \frac{9(\eta_{\lambda}^0)^2\ell^2\kappa^2 J}{\mu_2 \eta_0} \leq \frac{\mu_2\eta_0}{8 J}.
 \end{align*}
The inequalities of $\phi_1, \phi_2$ can be simplified as
\begin{align}
 \phi_1 & \leq \xi_k^1 \left(1 - \frac{53\mu_2\eta_0}{72 J K^{(a-c)}}\right) - \xi_k^1  + \frac{\ell^2\eta_{\lambda}^0}{K^{(b-2c)}} + \xi_k^2\frac{\mu_2 \eta_0}{9J K^{(a-c)}} \leq 0 \label{inequ:phi:1} \\
 \phi_2 &  \leq \xi_k^2 \left(1 - \frac{17\mu_2\eta_0}{72J K^{(a-c)}}\right) - \xi_k^2 + \frac{2\ell^2\eta_{\lambda}^0}{K^{(b-2c)}} + \xi_k^1\frac{\mu_2\eta_0}{9J K^{(a-c)}} \leq 0 \label{inequ:phi:2}
\end{align}
We might solve the above inequalities and properly set 
 \begin{align*}
    \xi_k^1  & = \frac{-\frac{53\mu_2\eta_0}{72 J K^{(a-c)}}\frac{2\ell^2\eta_{\lambda}^0}{K^{(b-2c)}} - \frac{\ell^2\eta_{\lambda}^0}{K^{(b-2c)}}\frac{\mu_2\eta_0}{9J K^{(a-c)}}}{\frac{\mu_2 \eta_0}{9J K^{(a-c)}}\frac{\mu_2\eta_0}{9J K^{(a-c)}} - \frac{17\mu_2\eta_0}{72J K^{(a-c)}} \frac{53\mu_2\eta_0}{72J K^{(a-c)}}} = \frac{\frac{114\ell^2\eta_{\lambda}^0 }{ K^{(b-2c)}}}{\frac{837\mu_2\eta_0}{72JK^{(a-c)}}} = \frac{10\ell^2\eta_{\lambda}^0}{\mu_2\eta_0} \frac{J}{K^{(b-a-c)}}\notag \\
    \xi_k^2  &  = \frac{-\frac{17\mu_2\eta_0}{72 J K^{(a-c)}}\frac{\ell^2\eta_{\lambda}^0}{K^{(b-2c)}} - \frac{2\ell^2\eta_{\lambda}^0}{K^{(b-2c)}}\frac{\mu_2\eta_0}{9J K^{(a-c)}}}{\frac{\mu_2 \eta_0}{9J K^{(a-c)}}\frac{\mu_2\eta_0}{9J K^{(a-c)}} - \frac{17\mu_2\eta_0}{72J K^{(a-c)}} \frac{53\mu_2\eta_0}{72J K^{(a-c)}}}  = \frac{3\ell^2\eta_{\lambda}^0}{\mu_2\eta_0} \frac{J}{K^{(b-a-c)}}
 \end{align*}
 to guarantee that $\phi_1 \leq 0$ and $\phi_2 \leq 0$. Then the main inequality (\ref{inequ:main:lyapu}) can be estimated as
 \begin{align*}
  \E[R_{k+1} \mid \mathcal{F}_k, \mathcal{C}_k] 
& \leq R_k - \left(\frac{\eta_{\lambda}}{2} - \ell_{\Gamma}  \eta_{\lambda}^2 
 - 2\xi_{k+1}^1 (1+\gamma_1)\kappa^2\eta_{\lambda}^2 - 2\xi_{k+1}^2 (1+\gamma_2)\kappa^2\eta_{\lambda}^2\right) \left\| \nabla \Gamma^{\alpha}(\lambda_k) \right\|^2  \notag \\
 & + \frac{\ell_{\Gamma}  \eta_{\lambda}^2}{2}\left(\sigma_1^2 + 2\alpha^2\sigma_2^2\right)  + \xi_{k+1}^1(1+\gamma_1^{-1})\frac{\alpha^2 \eta_{u}^2\sigma_2^2}{JB} + \xi_{k+1}^2(1+\gamma_2^{-1})\frac{\eta_{w}^2\left(\sigma_1^2 + \alpha^2 \sigma_2^2\right)}{JB}.
 \end{align*}
If we set $\eta_{\lambda}^0 \leq 1/(8\ell_{\Gamma})$, then $\ell_{\Gamma}  \eta_{\lambda}^2 \leq \frac{\eta_{\lambda}}{8}$. For $b \geq a$ and $ k \geq 1$, if we set $\eta_0 / \eta_{\lambda}^0 \geq 8\sqrt{3}\kappa^2 J$
\begin{align*}
\xi_k^1(1+\gamma_1)\kappa^2\eta_{\lambda} & \leq  \frac{40(\eta_{\lambda}^0)^2 K^{(a-c)}\ell^2 \kappa^2  J^2}{\mu_2^2\eta_0^2K^{b}K^{(b-a-c)}} = \frac{40(\eta_{\lambda}^0)^2 \ell^2 \kappa^2  J^2}{\mu_2^2\eta_0^2K^{(2b-2a)}}\leq \frac{1}{16} \\
\xi_k^2(1+\gamma_2)\kappa^2\eta_{\lambda} & \leq \frac{12(\eta_{\lambda}^0)^2 K^{(a-c)}\ell^2 \kappa^2  J^2}{\mu_2^2\eta_0^2K^{b}K^{(b-a-c)}} = \frac{12(\eta_{\lambda}^0)^2 \ell^2 \kappa^2  J^2}{\mu_2^2\eta_0^2K^{(2b-2a)}} \leq \frac{1}{16}.
\end{align*}
Then
\begin{align*}
     \E[R_{k+1} \mid \mathcal{F}_k, \mathcal{C}_k] 
& \leq R_k -  \frac{\eta_{\lambda}}{4}  \left\| \nabla \Gamma^{\alpha}(\lambda_k) \right\|^2  + \frac{\ell_{\Gamma}  \eta_{\lambda}^2}{2}\left(\sigma_1^2 + 2\alpha^2\sigma_2^2\right)  + \frac{8\xi_{k+1}^1}{7}\frac{\alpha^2 \eta_{u}^2\sigma_2^2}{JB} + \frac{3\xi_{k+1}^2}{4}\frac{\eta_{w}^2\left(\sigma_1^2 + \alpha^2 \sigma_2^2\right)}{JB}.
\end{align*}
Telescoping the above inequality gives
\begin{align*}
 \E\left[\left\|\nabla \Gamma^{\alpha}(\tilde{\lambda})  \right\|^2 \right] & = \frac{1}{K}\sum_{k=1}^{K} \E\left[\left\| \nabla \Gamma^{\alpha}(\lambda_k) \right\|^2\right]  \notag \\
 & \leq \frac{4}{K\eta_{\lambda}} \left(\sum_{k=1}^{T} \E[R_k \mid \mathcal{F}_{k-1}, \mathcal{C}_{k-1}] - \E[R_{k+1} \mid \mathcal{F}_k], \mathcal{C}_{k} \right) \notag \\
 & + \frac{4}{K\eta_{\lambda}}\sum_{k=1}^K \left(\frac{\ell_{\Gamma}  \eta_{\lambda}^2}{2}\left(\sigma_1^2 + 2\alpha^2\sigma_2^2\right)  + \frac{8\xi_{k+1}^1}{7}\frac{\alpha^2 \eta_{u}^2\sigma_2^2}{JB} + \frac{3\xi_{k+1}^2}{4}\frac{\eta_{w}^2\left(\sigma_1^2 + \alpha^2 \sigma_2^2\right)}{JB} \right) \notag \\
& \leq  \frac{4 \E[R_1]  K^{b}}{\eta_{\lambda}^0 K} + \frac{4K^{b}}{\eta_{\lambda}^0} \frac{\ell_{\Gamma} (\eta_{\lambda}^0)^2\left(\sigma_1^2 + K^{2c} \sigma_2^2\right)}{2K^{2b}} \notag \\
& + \frac{4K^{b}}{\eta_{\lambda}^0}\left( \frac{80\ell^2\eta_0\eta_{\lambda}^0 K^{2c} K^{-2a}\sigma_2^2}{7\mu_2B K^{(b-a-c)}} + \frac{9\ell^2\eta_0\eta_{\lambda}^0 K^{-2a}\left(\sigma_1^2 +  K^{2c}\sigma_2^2\right)}{4\mu_2B K^{(b-a-c)}}\right).
\end{align*}
Recalling the result of Lemma~\ref{lem:equiv} states the relation between the stationarity of the minimax problem and the original bilevel problem, we have
\begin{align*}
& \E\left[\left\|\nabla \cL(\tilde{\lambda})  \right\|^2 \right] = \frac{1}{K}\sum_{k=1}^{K} \E\left[\left\| \nabla \cL(\lambda_k) \right\|^2\right] \notag \\
 & \leq  \frac{2}{K}\sum_{k=1}^{K} \left(\E\left[\left\| \nabla \cL(\lambda_k) - \nabla \Gamma^{\alpha}(\lambda_k) \right\|^2\right] + \E\left[\left\| \nabla \Gamma^{\alpha}(\lambda_k) \right\|^2\right] \right) \notag \\
 & \leq \frac{2}{\alpha^2} + \frac{2}{K}\sum_{k=1}^{K}  \E\left[\left\| \nabla \Gamma^{\alpha}(\lambda_k) \right\|^2\right] \notag \\
& \leq  \frac{2}{K^{2c}} + \frac{8 \E[R_1]  K^{b}}{\eta_{\lambda}^0 K} + \frac{8K^{b}}{\eta_{\lambda}^0}\frac{\ell_{\Gamma} (\eta_{\lambda}^0)^2\left(\sigma_1^2 + K^{2c} \sigma_2^2\right)}{2K^{2b}} \notag \\
& + \frac{8K^{b}}{\eta_{\lambda}^0} \left(\frac{80\ell^2\eta_0\eta_{\lambda}^0 K^{2c} K^{-2a}\sigma_2^2}{7\mu_2B K^{(b-a-c)}} + \frac{9\ell^2\eta_0\eta_{\lambda}^0 K^{-2a}\left(\sigma_1^2 +  K^{2c}\sigma_2^2\right)}{4\mu_2B K^{(b-a-c)}}\right).
\end{align*}
Let $c = 1/7$, $a=4/7$, and $b=5/7$, we have
\begin{align*}
 \E\left[\left\|\nabla \cL(\tilde{\lambda})  \right\|^2 \right] &= \frac{1}{K}\sum_{k=1}^{K} \E\left[\left\| \nabla \cL(\lambda_k) \right\|^2\right] \notag \\
& \leq \mathcal{O}\left(\frac{1}{K^{2/7}}\right)+  \mathcal{O}\left(\frac{\E[R_1]}{\eta_{\lambda}^0K^{2/7}}\right) +  \mathcal{O}\left(\frac{(1+\ell \kappa \eta_0)\sigma_1^2}{B K^{4/7}} \right) + \mathcal{O}\left(\frac{(1+\ell \kappa \eta_0)\sigma_2^2}{B K^{2/7}} \right).
\end{align*}
Note that the initial state $R_1$ can be controlled by a constant which is independent with $\alpha$:
\begin{align}
R_1 & =  \Gamma^{\alpha}(\lambda_{1}) - \Gamma_{\min}^{\alpha} + \xi_1^{1}  \delta_{1} + \xi_{2}^1  r_{1} \notag \\
& = \Gamma^{\alpha}(\lambda_{1}) - \Gamma_{\min}^{\alpha} + \mathcal{O}\left(J \kappa \eta_0^{\lambda}/\eta_0 \left(\left\|w_1 - w_{\ast}^{\alpha}(\lambda_1) \right\|^2 + \left\|u_1 - u_{\ast}(\lambda_1)\right\|^2\right) \right)
\end{align}
where 
\begin{align}
\Gamma^{\alpha}(\lambda_{1}) - \Gamma_{\min}^{\alpha} &\leq \cL^{\alpha}(\lambda_1, w_{\ast}^{\alpha}(\lambda_1), u_{\ast}(\lambda_1)) - \cL^{\alpha}(\lambda_{\ast}, w_{\ast}^{\alpha}(\lambda_{\ast}), u_{\ast}(\lambda_{\ast})) \notag \\
 & = L_1(\lambda_1, w_{\ast}^{\alpha}(\lambda_1)) - L_1(\lambda^{\ast}, w_{\ast}^{\alpha}(\lambda^{\ast})) + \alpha \left(L_2(\lambda_1, w_{\ast}^{\alpha}(\lambda_1)) - L_2(\lambda_1, u_{\ast}(\lambda_1)) \right) \notag \\
& + \alpha \left(L_2(\lambda^{\ast}, w_{\ast}^{\alpha}(\lambda^{\ast})) - L_2(\lambda^{\ast}, u_{\ast}(\lambda^{\ast})) \right) \notag \\
 & =  L_1(\lambda_1, w_{\ast}(\lambda_1)) - L_1(\lambda^{\ast}, w_{\ast}(\lambda^{\ast})) + L_1(\lambda_1, w_{\ast}^{\alpha}(\lambda_1)) - L_1(\lambda_1, w_{\ast}(\lambda_1)) \notag \\ & + L_1(\lambda^{\ast}, w_{\ast}(\lambda^{\ast})) - L_1(\lambda^{\ast}, w_{\ast}^{\alpha}(\lambda^{\ast})) + \alpha \left(L_2(\lambda_1, w_{\ast}^{\alpha}(\lambda_1)) - L_2(\lambda_1, u_{\ast}(\lambda_1)) \right) \notag \\
& + \alpha \left(L_2(\lambda^{\ast}, w_{\ast}^{\alpha}(\lambda^{\ast})) - L_2(\lambda^{\ast}, u_{\ast}(\lambda^{\ast})) \right) \notag \\
\leq & \cL(\lambda_1) - \cL(\lambda^{\ast}) + \ell_{10}\left\| w_{\ast}^{\alpha}(\lambda_1) - w_{\ast}(\lambda_1)\right\| + \ell_{10}\left\| w_{\ast}^{\alpha}(\lambda^{\ast}) - w_{\ast}(\lambda^{\ast})\right\| \notag \\
& + \alpha \frac{\ell_{21}}{2}\left\|w_{\ast}^{\alpha}(\lambda_1)-u_{\ast}(\lambda_1)\right\|^2 +  \alpha \frac{\ell_{21}}{2}\left\|w_{\ast}^{\alpha}(\lambda^{\ast})-u_{\ast}(\lambda^{\ast})\right\|^2 \notag \\
 & \leq \cL(\lambda_1) - \cL(\lambda^{\ast}) + \frac{2\ell_{10}C_0}{\alpha}  + 2\alpha \frac{\ell_{21}}{2} \frac{C_0^2}{\alpha^2} \notag \\
 & \leq \cL(\lambda_1) - \cL(\lambda^{\ast}) + \frac{2\ell_{10}C_0 \mu_2}{\ell_{11}}  + \frac{\ell_{21} C_0^2\mu_2}{\ell_{11}} = \cL(\lambda_1) - \cL(\lambda^{\ast}) + \cO\left( \kappa^2 \ell_{21}\right),
\end{align}
where by definitions we know $w_{\ast}(\lambda) = u_{\ast}(\lambda)$ and the first inequality follows from the gradient-Lipschitz of $L_2$ and the Lipschitz continuity of $L_1$ in $w$, and the second inequality uses Lemma~\ref{lem:y:ast}. The proof is complete.
\end{proof}

\end{document}